\pdfoutput=1

\documentclass[11pt]{article}

\usepackage[final]{acl}

\usepackage{times}
\usepackage{latexsym}

\usepackage[T1]{fontenc}

\usepackage[utf8]{inputenc}

\usepackage{microtype}

\usepackage{inconsolata}

\usepackage{inconsolata}
\usepackage{amsmath} 
\usepackage{pifont}
\usepackage{hyperref}
\usepackage{arydshln}
\usepackage{booktabs}
\usepackage{array}
\usepackage{multirow} 
\usepackage{float}
\usepackage{graphicx} 
\usepackage{makecell}
\usepackage{algorithm}
\usepackage{algpseudocode}
\usepackage{comment}
\usepackage{CJKutf8}
\usepackage[table]{xcolor}
\usepackage[dvipsnames]{xcolor}
\usepackage{colortbl}
\usepackage{tabularx}
\usepackage{relsize}

\usepackage{enumitem}

\title{\textsc{HiMATE}: A Hierarchical Multi-Agent Framework for Machine Translation Evaluation}

\usepackage[misc]{ifsym}
\usepackage{footmisc}
\DefineFNsymbols*{newfootnote}{%
    \textasteriskcentered\dag{\textrm{\Letter}}\textdaggerdbl{\ding{73}}\P{**}%
    {\ding{172}}{\ding{173}}{\ding{174}}}
\setfnsymbol{newfootnote}

\newcommand*\samethanks[1][\value{footnote}]{\footnotemark[#1]}
\author{
Shijie Zhang$^{1}$\thanks{Equal contribution.} \and
Renhao Li$^{1,2}$\samethanks[1]\thanks{Under the Joint Ph.D. Program between UM and SIAT.} \and
Songsheng Wang$^1$\samethanks[1] \and \\
\bf{Philipp Koehn}$^3$ \and
\bf{Min Yang}$^2$\thanks{Corresponding author.} \and
\bf{Derek F. Wong}$^1$\samethanks[3] \\
$^1$ NLP$^2$CT Lab, Department of Computer and Information Science, University of Macau \\
$^2$ Shenzhen Key Laboratory for High Performance Data Mining, \\Shenzhen Institutes of Advanced Technology, Chinese Academy of Sciences \\
$^3$ Johns Hopkins University \\
\texttt{nlp2ct.\{shijie,songsheng\}@gamil.com,
li.renhao@connect.um.edu.mo}\\
\texttt{phi@jhu.edu,
min.yang@siat.ac.cn,
derekfw@um.edu.mo}
}

\begin{document}
\newcommand{\framework}{HiMATE}
\newcommand{\nlg}{Nature Language Generation}
\newcommand{\llm}{Large Language model}
\newcommand{\bleu}{BLEU}
\newcommand{\bertscore}{BERTSCORE}
\newcommand{\comet}{COMET-22}
\newcommand{\cometkiwi}{COMETKIWI}
\newcommand{\xcomet}{xCOMET-XL}
\newcommand{\gembamqm}{GEMBA-MQM}
\newcommand{\mmad}{M-MAD}
\newcommand{\slice}[2]{{#1[\,#2\,]}}  
\newcommand{\stext}{\ \text{such that}\ }  

\setlength{\dashlinedash}{2pt}    
\setlength{\dashlinegap}{1.5pt}   
\setlength{\hdashlinewidth}{0.4pt}

\maketitle
\begin{abstract}
The advancement of Large Language Models~(LLMs) enables flexible and interpretable automatic evaluations. In the field of machine translation evaluation, utilizing LLMs with translation error annotations based on Multidimensional Quality Metrics~(MQM) yields more human-aligned judgments. However, current LLM-based evaluation methods still face challenges in accurately identifying error spans and assessing their severity.
In this paper, we propose {\framework}, a \textbf{Hi}erarchical \textbf{M}ulti-\textbf{A}gent Framework for Machine \textbf{T}ranslation \textbf{E}valuation. We argue that existing approaches inadequately exploit the fine-grained structural and semantic information within the MQM hierarchy. To address this, we develop a hierarchical multi-agent system grounded in the MQM error typology, enabling granular evaluation of subtype errors. Two key strategies are incorporated to further mitigate systemic hallucinations within the framework: the utilization of the model's self-reflection capability and the facilitation of agent discussion involving asymmetric information.
Empirically, {\framework} outperforms competitive baselines across different datasets in conducting human-aligned evaluations. Further analyses underscore its significant advantage in error span detection and severity assessment, achieving an average F1-score improvement of 89\% over the best-performing baseline. We make our code and data publicly available at \url{https://github.com/nlp2ct-shijie/HiMATE}.
\end{abstract}

\section{Introduction}
Translation capability reflects the cross-lingual comprehension and generation quality of Large Language Models~(LLMs)~\cite{hendy2023good,yang2024qwen2,dubey2024llama}. Compared to traditional machine translation models, applying LLMs significantly enhances the translation capabilities of machine translation systems and introduces new challenges to Machine Translation Evaluation~(MTE)~\cite{kocmi-etal-2024-findings}.

\begin{figure}[t]
    \centering
    \includegraphics[width=1\linewidth]{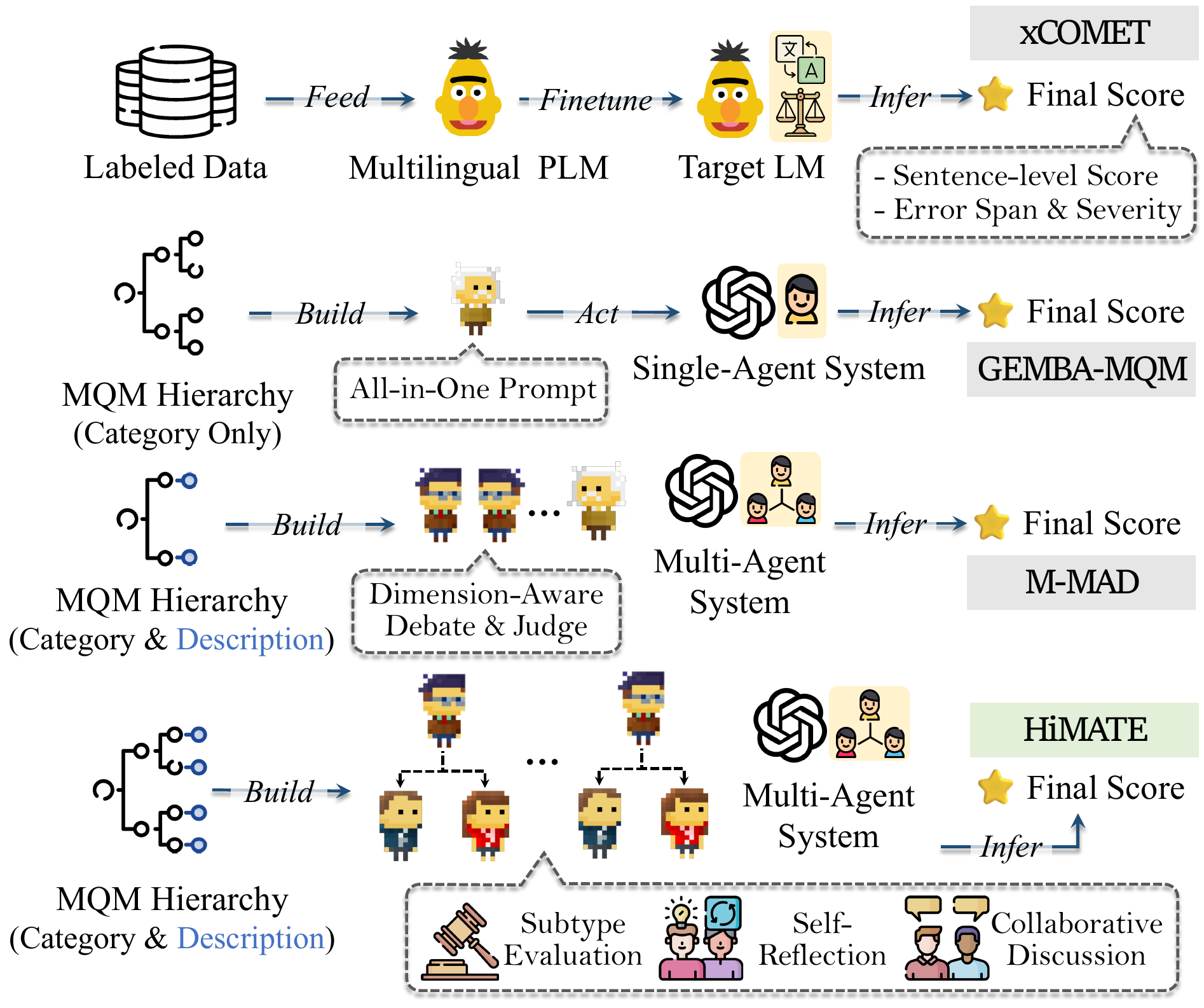}
    \caption{A comparison of \textsc{{\framework}} and other representative MQM-based metrics for MTE.}
    \label{fig:intro}
\end{figure}

Previously, statistics-based MTE metrics such as BLEU~\cite{papineni-etal-2002-bleu} and METEOR~\cite{banerjee-lavie-2005-meteor} are utilized for their efficiency, despite inherent limitations in accurately capturing semantic similarity. To address this gap, learned metrics such as COMET-20~\cite{rei-etal-2020-comet} and RoBLEURT~\cite{wan-etal-2021-robleurt} have been developed to enhance evaluations. 
These neural metrics can further improve performance via test-time adaptation by uncertainty minimization~\citep{zhan-etal-2023-test}. In addition, there are methods to extend the evaluation dimensions to translation difficulty~\citep{zhan-etal-2021-difficulty} and the fuzziness of reference~\citep{zhang2023obscurity}.
However, these approaches primarily yield numeric scores and cannot identify the specific error spans in translations~\cite{fernandes-etal-2023-devil, leiter2022towards}.

The machine translation data, annotated based on the guidelines of Multidimensional Quality Metrics~(MQM),\footnote{\href{https://themqm.org/}{themqm.org}.} includes detailed information regarding error spans, categories, and severity for each translated sentence, which facilitates more precise finetuning and thorough analysis of MTE models. 
The xCOMET model~\cite{guerreiro-etal-2024-xcomet} is finetuned on such data for interpretable scores, while it necessitates resource-intensive multi-stage training and shows a lack of generalization ability. The LLM-based single-agent evaluation systems, including EAPrompt~\cite{lu-etal-2024-error} and {\gembamqm}~\cite{kocmi-federmann-2023-gemba}, utilize MQM-informed prompts to guide LLMs in error detection. Nevertheless, the hallucination and position bias inherent within LLM evaluators affect their performance~\cite{wang-etal-2024-large-language-models-fair}.
In response to these challenges, the recently proposed method {\mmad}~\cite{feng2024m} employs a multi-agent system to decompose complex MTE tasks into simpler assessments, showcasing the potential of this paradigm. However, its effectiveness primarily stems from the self-consistency of agents operating within identical contexts, which may limit nuanced, multi-level error analysis. We argue that the rich semantic and hierarchical structure of the MQM framework has been largely underutilized in current methodological designs. By developing a more fine-grained, hierarchical agent-based system explicitly tailored to leverage the untapped potential of MQM’s structured hierarchy, error detection accuracy could be significantly enhanced, thereby narrowing the gap with human expert performance.

In this paper, we propose \textsc{{\framework}}, a \textbf{Hi}erarchical \textbf{M}ulti-\textbf{A}gent Framework for Machine \textbf{T}ranslation \textbf{E}valuation. Specifically, we establish the agent topology based on the MQM error typology, facilitating hierarchical agent nodes to exchange information in alignment with human error classification methodology. Error descriptions associated with each node are then employed to initialize multi-agent evaluators, allowing them to focus on detailed, simplified discrimination tasks rather than evaluations against broad, comprehensive criteria. This design facilitates fine-grained subtype error evaluations through multi-agent collaboration. Furthermore, we propose two post-stages to enhance the accuracy of the system in error detection and severity assessment.
Concretely, the subtype evaluator first revises the original translations based on its error detection results, then re-evaluates the validity of the initial judgments through comparison. Cases with low confidence during this self-reflection phase are forwarded to the next stage, where agents from different tiers collaboratively determine the presence and severity of translation errors through iterative discussion.
In contrast to prior methods, the proposed method leverages MQM information more effectively for decomposing complex tasks and constructing hierarchical multi-agent systems. By guiding these agents through self-reflection and collaboration, we achieve more human-aligned automatic evaluations. Figure~\ref{fig:intro} highlights the key differences between {\framework} and other representative MTE metrics.

We summarize our contributions as follows:
\begin{itemize}[leftmargin=*]
    \item[\textbullet] We propose \textsc{{\framework}}, a novel multi-agent machine translation evaluation framework that leverages MQM hierarchy-derived structural and semantic information to define agent topology, resulting in reliable, human-aligned judgments.
    \item[\textbullet] We develop a three-stage strategy to enhance error detection and severity assessment within the framework, facilitated by self-reflection and collaborative discussions among agents.
    \item[\textbullet] Empirical results across various models and datasets demonstrate the effectiveness of \textsc{{\framework}}. Further analyses highlight its superiority in both error span identification and severity evaluation, achieving an average 89\% improvement in F1-score and 95\% enhancement in Recall over the best-performing baseline.
\end{itemize}

\section{Related Work}
\subsection{Machine Translation Evaluation}
The objective of MTE is to assess the quality of machine-translated text. Traditional statistics-based metrics~\cite{papineni-etal-2002-bleu,lin-2004-rouge,zhang2019bertscore} often inadequately capture semantic similarity~\cite{rei-etal-2020-comet}. Certain learned metrics~\cite{rei-etal-2022-comet,rei-etal-2022-cometkiwi,guerreiro-etal-2024-xcomet} have shown improvement, while high iteration costs and limited generalization ability constrain these methods. 
UniTE~\citep{wan-etal-2022-unite} improves the performance of various translation tasks by using monotonic regional attention to control the interaction among input segments. BARTScore++~\citep{lu-etal-2023-toward} enhances BARTScore~\citep{yuan2021bartscore} by introducing human-like error analysis strategies.
Recent advancements harness the power of LLMs for translation evaluation. GEMBA-DA~\citep{kocmi-federmann-2023-large} implements direct scoring through prompt tuning, whereas EAPrompt~\citep{lu-etal-2024-error} identifies errors without numerical scoring to mitigate bias. AUTOMQM~\citep{fernandes-etal-2023-devil} and {\gembamqm}~\citep{kocmi-federmann-2023-gemba} prompt models to produce a score within the MQM framework. MQM-APE~\citep{lu-etal-2025-mqm} improves both the reliability and quality of error spans by introducing Automatically Post-Editing~(APE). However, these methods insufficiently utilize the hierarchical information of the MQM framework.

\begin{figure*}[!htb]
    \centering
    \includegraphics[width=1\linewidth]{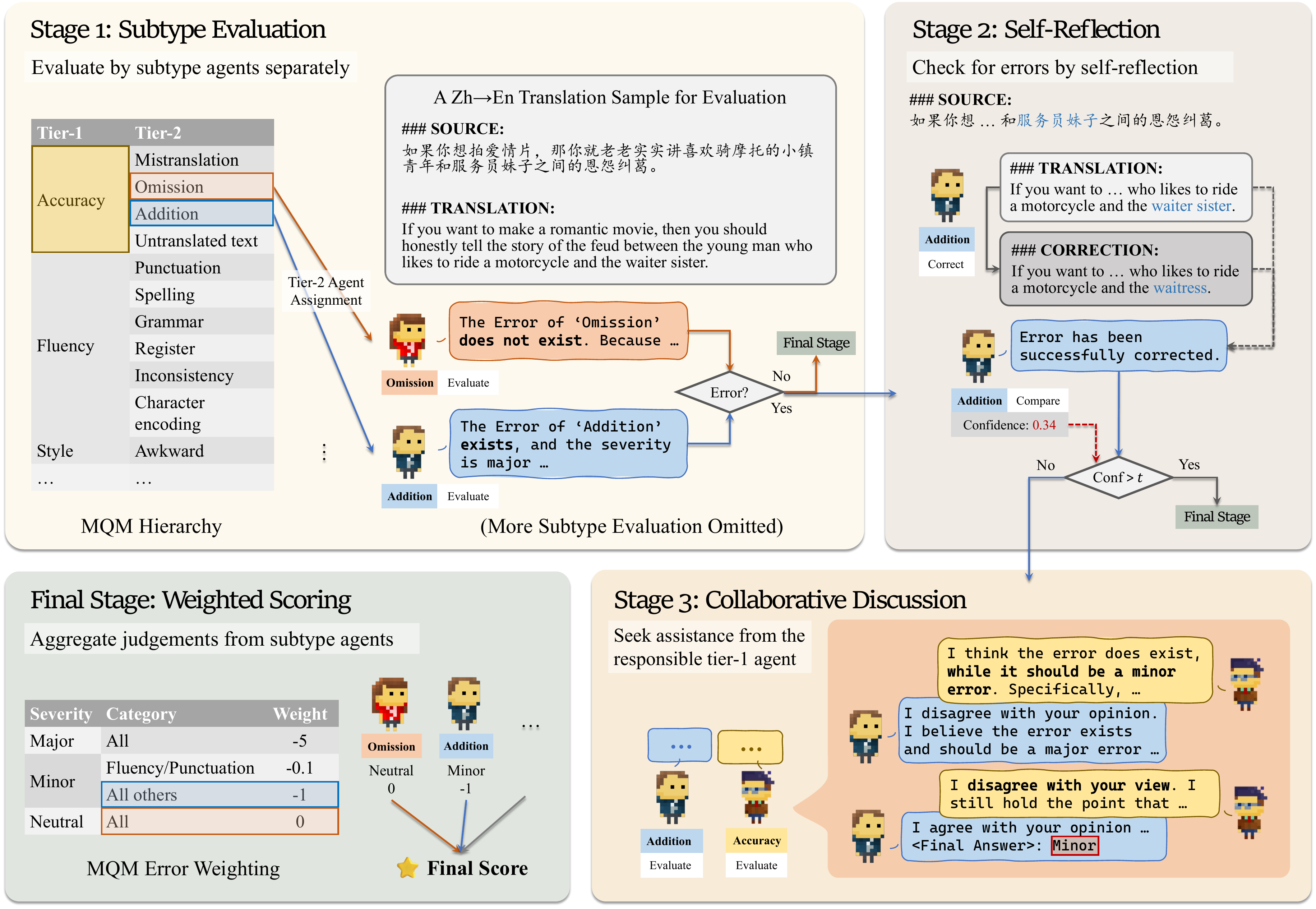}
    \caption{An overview of the proposed framework {\framework} where translation errors are categorized into tier-1 high-level errors and tier-2 fine-grained subtype errors. The tier-2 agent first evaluates a subtype error. Then the judgment is refined through the self-reflection stage. If needed, the tier-2 agent discusses with its upper-level tier-1 agent for further confirmation. The final score is computed by summing the weighted values of errors.}
    \label{fig:framework}
\end{figure*}

\subsection{Multi-Agent Evaluation}
Multi-agent systems have demonstrated their potential in enhancing the accuracy of evaluation within \nlg~(NLG). Chat\-Eval~\cite{chan2023chateval} employs LLMs in a framework similar to human group discussions. MAT\-Eval~\cite{li2024mateval} introduces architectures that rely on multi-round discussion and summarization for evaluation. Similarly, a courtroom-inspired architecture has been proposed, utilizing LLMs as adversarial judges engaging in multi-round debates for NLG assessment~\cite{bandi2024adversarial}. M-MAD~\cite{feng2024m} integrates multi-agent systems into MTE, leveraging the collaborative reasoning capabilities of LLMs. Nevertheless, existing approaches for MTE, such as M-MAD, insufficiently leverage fine-grained error definitions and semantic information within the MQM hierarchy.

\section{Methodology}
In this section, we introduce our proposed framework {\framework} for conducting human-aligned machine translation evaluations. Adhering to the hierarchical structure of the MQM error typology, {\framework} organizes translation errors into two tiers: high-level error categories~(tier-1) and fine-grained error subtypes~(tier-2). Two distinct types of agents are assigned to manage these tiers, with each agent specializing in assessing specific error categories, as depicted in Figure~\ref{fig:framework}. Initially, tier-2 agents individually perform evaluations on their corresponding error subtypes. This preliminary assessment is then refined by harnessing the self-reflection capabilities of LLMs. Finally, results exhibiting low confidence are subjected to collaborative discussion involving tier-1 and tier-2 agents. The entire evaluation process operates without reliance on reference, ensuring broad applicability for evaluation.

\subsection{Agent Topology}

The hierarchical error structure of the evaluation framework comprises high-level core error types $e_{i}$~(e.g., Accuracy), which are allocated to tier-1 agents $A_{i}^{[1]}$, and fine-grained error subtypes $e_{j}$~(e.g., Omission, Addition) that belong under these core errors and are assigned to tier-2 agents $A_{j}^{[2]}$. Specific descriptions are provided in Appendix~\ref{appendix_mqm_hierarchy}.

\subsection{Subtype Evaluation}

Initially, the tier-2 agents are required to seek the specific errors individually within the translations. For a given source text $x$, translation $y$, and system prompt $p_{\text{SE}}$, each agent $A_{j}^{[2]}$ is provided with distinct error definitions corresponding to tier-2, enabling them to leverage detailed information effectively. All the agents adhere to a uniform definition of severity when evaluating identified errors.

Upon completion of this stage, a preliminary assessment concerning the error and the corresponding severity is produced. Evaluators also provide explanations supporting their judgments. The initial evaluation information $\text{I}_{\text{SE}}$ generated above is then passed on to the subsequent stage:

\begin{equation}
\text{I}_{\text{SE}} \leftarrow A_{j}^{[2]}(x, y, p_{\text{SE}})
\end{equation}

\noindent
If no errors are detected, results proceed directly to the final stage; otherwise, sentences flagged with errors advance to the subsequent stage.

\subsection{Self-Reflection}
To enhance the robustness of the preliminary evaluation, we further validate the initial assessment by harnessing the self-reflection capabilities of LLMs. The prompt $p_{\text{SR}}$ consists of two components, $p_{\text{c}}$ for error correction and $p_{\text{v}}$ for comparative verification. Initially, the tier-2 agents scrutinize all error annotations and justifications from the preliminary assessment. Utilizing the information $\text{I}_{\text{SE}}$, tier-2 agents generate the corrected versions $y_{\text{c}}$:

\begin{equation}
y_{\text{c}} \leftarrow A_{j}^{[2]}(x, y, p_{\text{c}},\text{I}_{\text{SE}})
\end{equation}  

Subsequently, the evaluators compare the original translation $y$ with its corrected version $y_{\text{c}}$. Suppose an error is accurately identified during the initial phase, the correction should either rectify the issue or mitigate its severity, thereby validating the initial judgment. Otherwise, this may indicate a false positive in the previous evaluation, suggesting that the purported error might not exist. This verification mechanism yields more precise error judgments, diminishing workload concerning error type and severity in subsequent stages. Following the stage, additional confirmation regarding error information $\text{I}_{\text{SR}}$ can be procured:

\begin{equation}
\text{I}_{\text{SR}} \leftarrow A_{j}^{[2]}(y, y_{\text{c}}, p_{\text{v}}),\text{I}_{\text{SE}}
\end{equation}

\noindent
Responses with confidence scores, calculated by summing token logarithmic probabilities, below a predefined stage-transition threshold advance to the subsequent phase, whereas those surpassing this threshold proceed directly to the final stage.

\subsection{Collaborative Discussion}

The final stage is designed to validate the ambiguous assessment through collaborative discussion. This methodology stems from the observation that judgments with low confidence often indicate potential inaccuracies requiring expert validation. 

In this phase, the tier-1 agents $A_{i}^{[1]}$, responsible for high-level error categorization, collaborate and engage in discussions with the tier-2 agents. These tier-1 agents possess comprehensive knowledge of all tier-2 subcategory error definitions within their respective domains, enabling them to evaluate translation from a broader perspective.

Throughout the discussion period, tier-1 agents first analyze the information, initially set as $\text{I}_{\text{SR}}$ at the start of the discussion, from tier-2 agents. Subsequently, tier-1 agents provide expert evaluations, integrated with the prior chat history to form an updated discussion record. Tier-2 agents then contribute their perspectives, indicating agreement or disagreement with the assessments provided by tier-1 agents, further updating the discussion record.

A prompt requiring agents to begin their responses with either ``I agree'' or ``I disagree'' is employed to determine whether the discussion should persist. The discussion continues iteratively until consensus is reached or the maximum number of discussion rounds is exceeded. The result that includes the type and severity of error can be derived to gain the corresponding scoring weight.

\begin{table*}[ht]
    \centering
    \renewcommand{\arraystretch}{1.15}
    \newcolumntype{M}[1]{>{\centering\arraybackslash}m{#1}}
    \caption{Results of MTE methods on the ZH-EN and EN-DE subsets of the MQM22 dataset. A higher Kendall's correlation coefficient~($\tau$) and Spearman's correlation coefficient~($s$), as well as a lower Mean Absolute Error~(MAE) and Mean Squared Error~(MSE), indicate better alignment with human evaluations. The abbreviations \textit{4o-mini} and \textit{qwen2.5} refer to the gpt-4o-mini and qwen2.5-72b-instruct models, respectively. Reference-based methods are indicated with a \colorbox{gray!30}{gray background}. The best result in each column is \textbf{bolded}, and the second-best is \underline{underlined}.}
    \begin{tabular*}{\textwidth}{@{\extracolsep{\fill}} M{1cm} l c c c c c c c c @{}}
        \toprule
        \multirow{2}{*}{Model} & \multirow{2}{*}{Method} & 
        \multicolumn{4}{c}{ZH-EN} & \multicolumn{4}{c}{EN-DE} \\ \cmidrule(lr){3-6} \cmidrule(lr){7-10}
        & & $\tau\uparrow$ & $s\uparrow$ & $MAE\downarrow$ & $MSE\downarrow$ & $\tau\uparrow$ & $s\uparrow$ & $MAE\downarrow$ & $MSE\downarrow$ \\ \midrule
        
        \rowcolor[gray]{0.9} 
        & {\bleu} & 0.172 & 0.233 & 0.6849 & 0.5064 & 0.161 & 0.213 & 0.6184 & 0.4322 \\ 
        \rowcolor[gray]{0.9} 
        & {\bertscore} & 0.300 & 0.400 & 0.3334 & 0.1349 & 0.208 & 0.275 & 0.3936 & 0.1970 \\ 
        \rowcolor[gray]{0.9} 
        & {\comet} & 0.369 & 0.489 & 0.1527 & 0.0306 & \underline{0.301} & \textbf{0.390} & 0.0948 & 0.0144 \\
        
        & {\cometkiwi} & 0.365 & 0.487 & 0.1821 & 0.0386 & 0.212 & 0.277 & 0.1351 & 0.0221 \\ 
        \cdashline{1-10}

        \multirow{3}*{\rotatebox[origin=c]{90}{4o-mini}}
        & {\gembamqm} & 0.381 & 0.464 & 0.0755 & 0.0131 & 0.270 & 0.316 & 0.0420 & 0.0052 \\ 
        & {\mmad} & 0.350 & 0.435 & 0.0379 & 0.0033 & 0.282 & 0.327 & \textbf{0.0164} & \textbf{0.0007} \\ 
        & {\framework} (ours)& 0.404 & 0.502 & 0.0421 & 0.0051 & 0.293 & 0.338 & 0.0250 & 0.0026 \\ 
        \cdashline{1-10}
        
        \multirow{3}*{\rotatebox[origin=c]{90}{qwen2.5}}
        & {\gembamqm} & 0.407 & 0.510 & 0.0419 & 0.0039 & 0.229 & 0.279 & 0.0428 & 0.0045 \\ 
        & {\mmad} & 0.297 & 0.365 & \underline{0.0323} & \textbf{0.0027} & 0.227 & 0.266 & 0.0286 & 0.0018 \\ 
        & {\framework} (ours)& 0.413 & 0.498 & \textbf{0.0281} & \underline{0.0028} & 0.248 & 0.274 & 0.0198 & 0.0016 \\ 
        \cdashline{1-10}
         
        \multirow{3}*{\rotatebox[origin=c]{90}{qwen-max}}
        & {\gembamqm} & \underline{0.417} & \underline{0.527} & 0.0873 & 0.0159 & 0.292 & 0.347 & 0.0471 & 0.0071 \\ 
        & {\mmad} & 0.350 & 0.425 & 0.0417 & 0.0050 & 0.216 & 0.253 & 0.0258 & 0.0017 \\ 
        & {\framework} (ours)& \textbf{0.425} & \textbf{0.531} & 0.0386 & 0.0045 & \textbf{0.318} & \underline{0.358} & \underline{0.0187} & \underline{0.0014} \\ \bottomrule
    \end{tabular*}
    
    \label{tab:main_result}
\end{table*}

\subsection{Weighted Scoring}

Commencing with evaluation on subtype errors, {\framework} refines judgments through a self-reflection process to validate identified errors, eventually engaging in a collaborative discussion to reach consensus. After these three stages, the evaluation outcome for the translation sentence pair~$(x,y)$ regarding subtype error $e_j$ is obtained, consisting of the severity label $s_j$ and its corresponding weight $w_j$. The weights associated with each error subtype and severity level are detailed in Appendix~\ref{appendix_mqm_weight}. The score $S$ can be calculated as:

\begin{equation}
\text{S} = -\sum_{} w_{\text{j}}
\end{equation}

\begin{table*}[ht]
    \centering
    \renewcommand{\arraystretch}{1.15}
    \newcolumntype{M}[1]{>{\centering\arraybackslash}m{#1}}
    \caption{The system-level meta-evaluation results of MTE methods on the EN-DE subsets of the MQM24 dataset. A higher meta score means a better alignment with manual evaluation. The abbreviations \textit{4o-mini} and \textit{qwen2.5} refer to the gpt-4o-mini and qwen2.5-32b-instruct models, respectively. The best result of meta score is \textbf{bolded}.}
    \begin{tabular*}{\textwidth}{@{\extracolsep{\fill}} M{1cm} l c c c c c @{}}
        \toprule
        \multirow{2}{*}{Model} & \multirow{2}{*}{Method} & \multirow{2}{*}{Meta} & system-level & system-level & segment-level & segment-level \\
         & & & Acc & Pearson & Acc-t & Pearson \\
        \midrule

         & {\xcomet} & \textbf{0.731} & 0.844 & 0.989 & 0.539 & 0.550 \\
        \cdashline{1-7}

        \multirow{3}*{\rotatebox[origin=c]{90}{4o-mini}}
        & {\gembamqm} & 0.668 & 0.806 & 0.946 & 0.505 & 0.416 \\
        & {\mmad} & 0.656 & 0.750 & 0.889 & 0.500 & 0.485 \\
        & {\framework} (ours) & 0.708 & 0.861 & 0.968 & 0.533 & 0.469 \\
        \cdashline{1-7}
        
        \multirow{3}*{\rotatebox[origin=c]{90}{qwen2.5}}
        & {\gembamqm} & 0.692 & 0.844 & 0.997 & 0.523 & 0.404 \\
        & {\mmad} & 0.699 & 0.911 & 0.966 & 0.465 & 0.425 \\
        & {\framework} (ours) & 0.712 & 0.844 & 0.996 & 0.519 & 0.491 \\
        \cdashline{1-7}

        \multirow{3}*{\rotatebox[origin=c]{90}{qwen-max}}
        & {\gembamqm} & 0.726 & 0.889 & 0.998 & 0.545 & 0.471 \\
        & {\mmad} & 0.676 & 0.867 & 0.995 & 0.455 & 0.389 \\
        & {\framework} (ours) & \textbf{0.731} & 0.867 & 0.998 & 0.543 & 0.515 \\

        \bottomrule
        
    \end{tabular*}
    \label{tab:main_result_mqm24}
\end{table*}

\section{Experiments}

\begin{table*}[ht]
    \centering
    \renewcommand{\arraystretch}{1.13}  
    \newcolumntype{M}[1]{>{\centering\arraybackslash}m{#1}}
    \caption{Ablation study of {\framework} on the MQM22 dataset across different backbone models. The abbreviations \textit{4o-mini} and \textit{qwen2.5} refer to the gpt-4o-mini and qwen2.5-72b-instruct models, respectively. 
    SA denotes the single-agent baseline method. SE denotes the subtype evaluation stage. SR refers to the self-reflection stage. CD stands for the collaborative discussion stage.
    The best result for each metric within the same backbone model group is \textbf{bolded}.}
    \begin{tabular*}{\textwidth}{@{\extracolsep{\fill}} M{1cm} l c c c c c c c c @{}}
        \toprule
        \multirow{2}{*}{Model} & \multirow{2}{*}{Method} & 
        \multicolumn{4}{c}{ZH-EN} & \multicolumn{4}{c}{EN-DE} \\ \cmidrule(lr){3-6} \cmidrule(lr){7-10}
        & & $\tau\uparrow$ & $s\uparrow$ & $MAE\downarrow$ & $MSE\downarrow$ & $\tau\uparrow$ & $s\uparrow$ & $MAE\downarrow$ & $MSE\downarrow$ \\ 
        \midrule
        
        \multirow{5}*{\rotatebox[origin=c]{90}{4o-mini}}
        & SA & 0.345 & 0.417 & 0.0427 & \textbf{0.0042} & 0.210 & 0.230 & \textbf{0.0166} & \textbf{0.0008} \\ 
        \cdashline{2-10}
        & {\framework} &  &  &  &  &  &  &  &  \\ 
        & \hspace{0.5em}+SE & 0.359 & 0.468 & 0.1614 & 0.0614 & 0.278 & 0.329 & 0.0698 & 0.0205 \\ 
        & \hspace{0.5em}+SE+SR & 0.370 & 0.477 & 0.1178 & 0.0362 & 0.283 & 0.332 & 0.0497 & 0.0106 \\ 
        & \hspace{0.5em}+SE+SR+CD & \textbf{0.404} & \textbf{0.502} & \textbf{0.0421} & 0.0051 & \textbf{0.293} & \textbf{0.338} & 0.0250 & 0.0026 \\ 
        \midrule
        
        \multirow{5}*{\rotatebox[origin=c]{90}{\makecell[c]{qwen2.5}}}
        & SA & 0.341 & 0.413 & 0.0356 & 0.0031 & 0.183 & 0.200 & \textbf{0.0197} & 0.0017 \\ 
        \cdashline{2-10}
        & {\framework} &  &  &  &  &  &  &  &  \\ 
        & \hspace{0.5em}+SE & 0.397 & 0.489 & 0.0572 & 0.0172 & 0.240 & 0.267 & 0.0341 & 0.0100 \\ 
        & \hspace{0.5em}+SE+SR & 0.406 & 0.496 & 0.0479 & 0.0110 & 0.245 & 0.271 & 0.0281 & 0.0054 \\ 
        & \hspace{0.5em}+SE+SR+CD & \textbf{0.413} & \textbf{0.498} & \textbf{0.0281} & \textbf{0.0028} & \textbf{0.248} & \textbf{0.274} & 0.0198 & \textbf{0.0016} \\ 
        \midrule
        
        \multirow{5}*{\rotatebox[origin=c]{90}{qwen-max}}
        & SA & 0.395 & 0.480 & \textbf{0.0374} & 0.0049 & 0.286 & 0.325 & 0.0200 & \textbf{0.0014} \\ 
        \cdashline{2-10}
        & {\framework} &  &  &  &  &  &  &  &  \\ 
        & \hspace{0.5em}+SE & 0.409 & 0.523 & 0.1047 & 0.0405 & 0.292 & 0.333 & 0.0436 & 0.0140 \\ 
        & \hspace{0.5em}+SE+SR & 0.409 & 0.517 & 0.0925 & 0.0326 & 0.290 & 0.329 & 0.0354 & 0.0085 \\ 
        & \hspace{0.5em}+SE+SR+CD & \textbf{0.425} & \textbf{0.531} & 0.0386 & \textbf{0.0045} & \textbf{0.318} & \textbf{0.358} & \textbf{0.0187} & \textbf{0.0014} \\ 
        \bottomrule
    \end{tabular*}
    
    \label{tab:ablation_study}
\end{table*}

\subsection{Settings}
\noindent
\textbf{Dataset} \quad
We conduct experiments on the MQM22 dataset~\citep{freitag-etal-2022-results} and MQM24 dataset~\citep{freitag-etal-2024-llms}.
For the MQM22 dataset, we utilize the outputs from the HuaweiTSC system for Chinese-English~(ZH-EN) and the comet-bestmbr system for English-German~(EN-DE), encompassing 1,875 and 1,315 annotated translations respectively for segment-level evaluation.
Regarding the MQM24 dataset, we incorporate ten systems (including Aya23, Claude-3.5, CommandR-plus, Gemini-1.5-Pro, GPT-4, IKUN, IOL\_Research, Mistral-Large, MSLC, and TranssionMT) from the EN-DE dataset, which collectively contribute to a total of 4,870 samples, serving the purpose of system-level meta-evaluation.

\noindent
\textbf{Baseline} \quad
The reference-based evaluation metrics, including {\bleu}~\cite{papineni-etal-2002-bleu}, {\bertscore}~\cite{zhang2019bertscore}, and {\comet}~\cite{rei-etal-2022-comet}, as well as the reference-free ones, such as {\cometkiwi}~\cite{rei-etal-2022-cometkiwi}, {\xcomet}~\cite{guerreiro-etal-2024-xcomet}, {\gembamqm}~\cite{kocmi-federmann-2023-gemba}, and {\mmad}~\cite{feng2024m}, are adopted as baselines for comparison. Among these, {\gembamqm} and {\mmad} are LLM-based methods like ours.

\noindent
\textbf{Evaluation Method} \quad
For the segment-level evaluation, four widely adopted metrics are employed to evaluate the alignment of {\framework} with human judgments. Specifically, we use Kendall's correlation coefficient $\tau$ and Spearman's correlation coefficient $s$ to measure correlation, while using Mean Absolute Error~(MAE) and Mean Squared Error~(MSE) to quantify similarity. 
For system-level meta-evaluation, we use the same meta-evaluation metrics utilized in the WMT Metrics Shared Task~\citep{freitag-etal-2023-results}. The meta score is obtained by averaging four indicators: system-level pairwise accuracy, system-level Pearson correlation, segment-level Accuracy-t, and segment-level Pearson correlation.

\noindent
\textbf{Experiment Settings} \quad
The gpt-4o-mini~\cite{hurst2024gpt}, qwen2.5-72b-instruct, qwen2.5-32b-instruct, and qwen-max~\cite{yang2024qwen2} are adopted as backbone models for LLM-based MTE methods. The error typology used in {\framework} is categorized and further refined into 5 core error types and 19 distinct subtypes. The temperature parameter is set to 0. Several instances from the MQM20 and MQM21 datasets are selected to serve as 2-shot demonstration examples for the subtype evaluation stage. Stage-transition thresholds $t$ are adaptively configured for different models~(refer to Appendix~\ref{appendix_threshold} for specific procedures). The maximum number of dialogue turns during the collaborative discussion stage is set to four, with each agent allowed at most two statements. Prompts of {\framework} are shown in Appendix~\ref{appendix_prompts}.

\subsection{Main Results}
Table~\ref{tab:main_result} illustrates the experimental results of different MTE methods on the MQM22 dataset. 
Specifically, {\framework} showcases improved alignment with human evaluations, as evidenced by correlation coefficients. For ZH-EN, when utilizing the qwen-max model, {\framework} achieves peak correlation values of 0.425 in Kendall’s correlation coefficient and 0.531 in Spearman’s correlation coefficient. This advantage is sustained across diverse backbone models. Similar improvements are observed on the EN-DE dataset, where {\framework} surpasses other LLM-based methods, achieving the highest $\tau$ and second-best $s$.
Furthermore, {\framework} achieves either the best or second-best performance in MAE and MSE, showing comparable evaluation similarity with M-MAD and markedly exceeding other baseline methods. The significant improvement of similarity with human evaluation suggests the effectiveness of aligning human-built evaluation criteria within the proposed method. 

Table~\ref{tab:main_result_mqm24} reports the system-level meta-evaluation results on the MQM24 EN-DE dataset, which align with the findings from the segment-level evaluation. Among the LLM-based methods, {\framework} consistently outperforms alternatives, delivering higher meta scores across different backbone models. When using qwen-max as the backbone, our method attains a meta-evaluation score of 0.73, matching the performance of the specially trained {\xcomet}. These results further highlight the superior effectiveness of {\framework} compared with previous LLM-based metrics.

To sum up, {\framework} maintains strong performance across diverse backbone models, demonstrating robustness to variations in model capacity, and exhibits clear advantages at both segment and system levels. Collectively, these observations serve as compelling evidence of the efficacy of {\framework} in conducting high-quality machine translation evaluations.

\section{Ablation and Analysis}
\subsection{Ablation Study}
To verify the contribution of each stage in {\framework}, we conduct an ablation study in this section.
In addition to reporting the performance of individual stages, we also include the results of a single-agent evaluation method (denoted as SA) as a baseline ablation. Unlike the subtype evaluation stage of {\framework}, where each agent is dedicated to a specific error type, the SA baseline processes all tier-2 error definitions simultaneously and evaluates all types of errors at once. This baseline operates as a single stage and is capable of detecting multiple errors within a sentence.

Evaluation results of the ablated framework on the MQM22 dataset are shown in Table~\ref{tab:ablation_study}. According to the results, {\framework}-SE consistently outperforms SA in terms of correlation, particularly improving Kendall's correlation coefficient $\tau$ from 0.341 to 0.397 when using a qwen2.5 backbone model on the ZH-EN subset, which underscores the importance of fine-grained subtype error evaluation through a multi-agent approach.

The performance improvements from the self-reflection stage are relatively modest; however, this stage helps reduce the computational cost of the subsequent stage by filtering the high-confidence, reliable judgments. The collaborative discussion stage contributes more substantially, achieving the best or second-best results across all four evaluation metrics. It is worth noting that the performance improvements from collaborative discussion vary across different backbone LLMs, likely influenced by the instruction-following and divergent thinking capability of different models.


\begin{table*}[ht]
    \centering
    \caption{Evaluation results of different MTE methods on MQM22 ZH-EN under different sentence lengths. We use gpt-4o-mini as the backbone model. 
    Sentence separation thresholds have been set at 12 and 27 words.}
    \scalebox{0.95}{
    \begin{tabular*}{\textwidth}{@{\extracolsep{\fill}}lccccccccc@{}}
        \toprule
        & \multicolumn{3}{c}{short} & \multicolumn{3}{c}{medium} & \multicolumn{3}{c}{long} \\ 
        \cmidrule(lr){2-4} \cmidrule(lr){5-7} \cmidrule(lr){8-10}
        & $\tau$ & $s$ & F1-score & $\tau$ & $s$ & F1-score & $\tau$ & $s$ & F1-score \\ 
        \midrule
        {\gembamqm}  & \textbf{0.369} & 0.433 & 0.085 & 0.382 & 0.460  & 0.093 & 0.344 & 0.427 & 0.100   \\
        {\mmad}   & 0.333 & 0.386 & 0.124 & 0.316 & 0.392 & 0.108 & 0.311 & 0.408 & 0.116 \\
        {\framework} & 0.368 & \textbf{0.436} & \textbf{0.218} & \textbf{0.425} & \textbf{0.529} & \textbf{0.260} & \textbf{0.377} & \textbf{0.491} & \textbf{0.294} \\
        \bottomrule
    \end{tabular*}
    }
    \label{tab:long_text_analysis}
\end{table*}

\begin{figure*}[!t]
    \centering
    \includegraphics[width=1\linewidth]{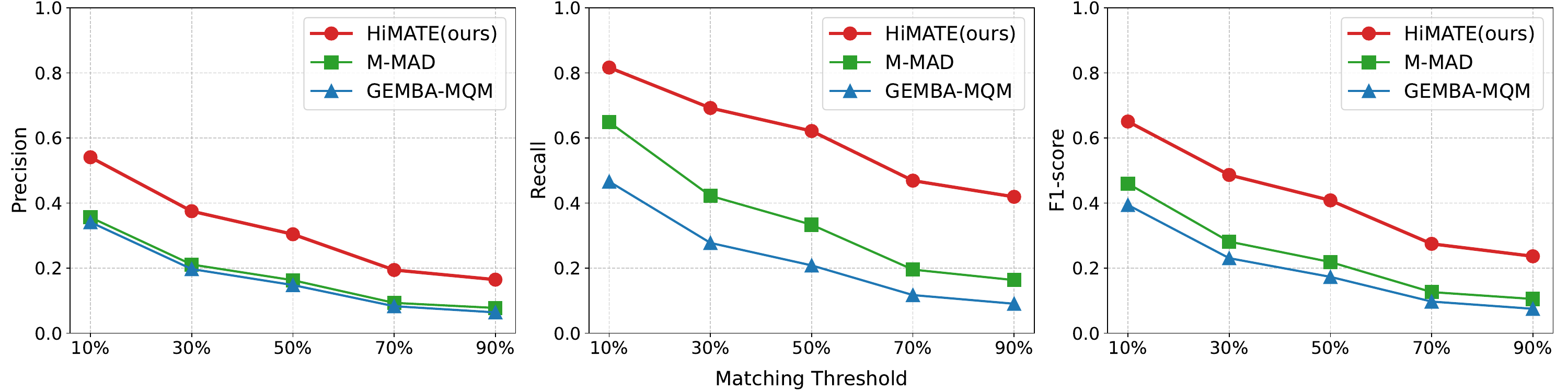}
    \caption{Error span detection results for different MTE methods on the MQM22 ZH-EN dataset, based on gpt-4o-mini. We report Precision, Recall, and F1-score across varying matching thresholds.}
    \label{fig:F1_score}
\end{figure*}

\begin{table}[t]
    \centering
    \small  
    \setlength{\tabcolsep}{3pt} 
    \caption{Domain-specific results of various MTE methods on MQM22 ZH-EN, measured by Spearman's correlation coefficient $s$ using gpt-4o-mini as backbone.}
    \begin{tabular}{@{} l *{4}{c} @{}}
        \toprule
          & news & conversation & social & ecommerce \\  
        \midrule
        {\gembamqm} & 0.333 & 0.367 & 0.471 & 0.538  \\ 
        {\mmad} & \textbf{0.365} & 0.347 & 0.430 & 0.444  \\ 
        {\framework} & 0.355 & \textbf{0.399} & \textbf{0.529} & \textbf{0.577}  \\ 
        \bottomrule
    \end{tabular}
    
    \label{tab:cross_domain_main}
\end{table}

\subsection{Error Span Detection}
As mentioned earlier, current LLM-based MTE methods struggle with accurately identifying translation error spans. In this section, we aim to evaluate the error span detection performance of {\framework} and several baseline approaches. Our experiments are conducted on the MQM22 ZH-EN subset, using gpt-4o-mini as the backbone model. Following~\citet{ye-etal-2024-mmad}, we utilize Precision, Recall, and F1-score for evaluation. Figure~\ref{fig:F1_score} illustrates the evaluation results at varying matching thresholds, which are defined by the minimum overlap ratio between detected error spans and gold-standard annotations. According to the results, {\framework} consistently surpasses {\gembamqm} and {\mmad} across all matching thresholds on all three metrics. Notably, compared to the best-performing baseline {\mmad}, {\framework} achieves average improvements of approximately 89\% in F1-score and 95\% in Recall, demonstrating its superior effectiveness in error span detection. These results further validate the effectiveness of {\framework}'s hierarchical agent design. More details regarding the error span matching algorithm are provided in Appendix~\ref{Appendix_error_span}.

\setlength{\dashlinedash}{2pt}    
\setlength{\dashlinegap}{1.5pt}   
\setlength{\hdashlinewidth}{0.4pt}

\begin{table*}[!ht]
    \centering
    \renewcommand{\arraystretch}{1.1}  
    \caption{A case of responses generated by gpt-4o-mini on MQM22 ZH-EN among different LLM-based methods. The parts marked in \textcolor{red}{red} indicate incorrect judgments, while the parts marked in \textcolor{green!50!black}{green} indicate correct judgments. 
    }
    \begin{tabular*}{\textwidth}{@{\extracolsep{\fill}}m{3cm} p{12.5cm}@{}}
        \toprule
        \multirow{1}{*}{\textbf{Source}} & 
        \begin{CJK*}{UTF8}{gbsn}希望你们了解一下
        \end{CJK*} \\
        \cdashline{1-2}
        \multirow{1}{*}{\textbf{Translation}} & 
        I hope you know about it. \\
        \cdashline{1-2}
        \multirow{1}{*}{\textbf{Reference}} & 
        I hope you can find out about it. \\
        \cdashline{1-2}
        \multirow{1}{*}{\textbf{Annotation}} & 
        Major-Mistranslation-‘know about it’ \\
        \midrule
        
        \multirow{2}{*}{\gembamqm} & \textcolor{green!50!black}{Major}-\textcolor{green!50!black}{Mistranslation}-‘I hope you \textcolor{green!50!black}{know about it}’, \textcolor{red}{Minor}-\textcolor{red}{Awkward}-‘I hope you \textcolor{green!50!black}{know about it}’ \\
        \cdashline{1-2}
        \multirow{1}{*}{{\mmad} Stage1} & \textcolor{red}{Minor}-\textcolor{red}{Omission}-‘\textcolor{green!50!black}{know about it}’\\
        \cdashline{1-2}
        \multirow{1}{*}{{\mmad} Stage2\&3} & \textcolor{red}{Minor}-\textcolor{red}{Omission}-‘\textcolor{green!50!black}{know about it}’ \\
        \cdashline{1-2}
        \multirow{3}{*}{{\framework}-SE} & \textcolor{red}{Minor}-\textcolor{red}{Inappropriate for context}-‘\textcolor{green!50!black}{know}’, \textcolor{red}{Major}-\textcolor{red}{Omission}-‘\textcolor{red}{\begin{CJK*}{UTF8}{gbsn}希望
        \end{CJK*}}’, \textcolor{red}{Minor}-\textcolor{red}{Awkward}-‘I hope you \textcolor{green!50!black}{know about it}’, \textcolor{red}{Minor}-\textcolor{red}{Addition}-‘\textcolor{green!50!black}{know about it}’, \textcolor{green!50!black}{Major}-\textcolor{green!50!black}{Mistranslation}-‘\textcolor{green!50!black}{know about it}’  \\
        \cdashline{1-2}
        \multirow{2}{*}{-SE+SR} & \textcolor{red}{Minor}-\textcolor{red}{Awkward}-‘I hope you \textcolor{green!50!black}{know about it}’, \textcolor{red}{Minor}-\textcolor{red}{Addition}-‘\textcolor{green!50!black}{know about it}’, \textcolor{green!50!black}{Major}-\textcolor{green!50!black}{Mistranslation}-‘\textcolor{green!50!black}{know about it}’ \\
        \cdashline{1-2}
        \multirow{1}{*}{-SE+SR+CD} & \textcolor{green!50!black}{Major}-\textcolor{green!50!black}{Mistranslation}-‘\textcolor{green!50!black}{know about it}’ \\
        \bottomrule
    \end{tabular*}
    
    \label{tab:case_study}
\end{table*}

\subsection{Performance on Distinct Sentence Length}
Longer contexts present significant challenges to both machine translation quality and evaluation accuracy. To investigate how sentence length affects MTE performance, we conduct experiments on the ZH-EN subset of MQM22.
We classify sentences containing more than 27 words as long, those ranging between 12 and 27 words as medium, and those comprising with fewer than 12 words as short.

As shown in Table~\ref{tab:long_text_analysis}, {\framework} consistently outperforms baselines across all sentence-length groups. While baseline methods achieve comparable results on shorter samples, {\framework} maintains superior performance as sentence length increases. In contrast, {\gembamqm} and {\mmad} exhibit limited or declining accuracy with rising complexity. Notably, at a matching threshold of 50\%, the F1-score gap between {\framework} and the best-performing baseline widens from 0.094 for short sentences to 0.194 for long sentences. These findings demonstrate {\framework}'s robustness in accurately identifying translation errors, particularly in longer sentences.

\subsection{Analysis on Domain-Specific Evaluation}

The MQM22 dataset encompasses four distinct text domains: news, social, conversation, and ecommerce. We present an analysis of the performance of different MTE methods across these domains using gpt-4o-mini as the backbone model. As illustrated in Table~\ref{tab:cross_domain_main}, {\framework} significantly surpasses the other two LLM-based methods in the conversation, social, and ecommerce domains. Although {\mmad} slightly outperforms {\framework} in the news domain, it exhibits the least effective performance in the other three domains. We also conduct experiments using other metrics and datasets, the details of which can be found in Appendix~\ref{appendix_cross_domain}. Overall, these findings underscore the robustness of {\framework} across diverse textual contexts.

\subsection{Case Study}
\label{case_study_main}
Through specific case studies, we compare the evaluation results of baseline methods with our proposed framework in Table~\ref{tab:case_study}, highlighting the distinct behavior of {\framework}. As shown, while {\gembamqm} correctly identifies the type and severity of one error, it inaccurately expands the error span to the entire sentence, making the assessment unreliable. Additionally, it falsely flags non-existent errors. Due to its single-agent, single-stage design, the system lacks the ability to refine or correct its judgments. In the case of {\mmad}, while the system precisely pinpoints the error span, it inaccurately classifies the error type and severity, and fails to rectify the previous misjudgment in the subsequent multi-agent debate stage. 
In contrast, while {\framework} initially over-identifies errors during the subtype evaluation stage, its self-reflection and collaborative discussion stages effectively filter out these incorrect assessments while retaining valid judgments. This demonstrates {\framework}'s effectiveness in allocating different tiers of hierarchical MQM information to distinct agents, enabling each to focus specifically on the error types under its responsibility, thereby enhancing the evaluation accuracy. More detailed case examples with staged evaluation processes across agents are provided in Appendix~\ref{appendix_case_study}.

\section{Conclusion}
This paper introduces {\framework}, a hierarchical multi-agent framework for LLM-based machine translation evaluation. The framework assembles a three-stage strategy to enhance error detection and severity assessment, leveraging self-reflection and collaborative discussion mechanisms after subtype evaluation. 
The experimental results of segment-level evaluation and system-level meta-evaluation demonstrate the superior performance of {\framework}.
Ablation study further validates the contribution of each stage in aligning closely with human judgments. Detailed analyses highlight that {\framework} exhibits pronounced advantages in error span detection compared to existing LLM-based methods. Additionally, evaluations across varying sentence lengths indicate robust and consistent performance. {\framework} also achieves stable advancements across various text domains. The achievement of {\framework} emphasizes the significance of effectively utilizing the human evaluation framework in designing multi-agent collaborative evaluation strategies.

\section*{Limitations}

Due to constraints in time and resources, the validation process presented in this study primarily focuses on selected models and language pairs, rather than an exhaustive evaluation across diverse models and languages. The current study does not evaluate the applicability of the proposed framework to advanced reasoning models, which have shown promising potential in comprehensive tasks. Additionally, the current experiment does not include the evaluation of the cross-lingual comprehension and generation quality of LLMs.

\section*{Acknowledgements}
This work was supported in part by the Science and Technology Development Fund of Macau SAR (Grant No. FDCT/0007/2024/AKP), the Science and Technology Development Fund of Macau SAR (Grant No. FDCT/0070/2022/AMJ, China Strategic Scientific and Technological Innovation Cooperation Project Grant No. 2022YFE0204900), the Science and Technology Development Fund of Macau SAR (Grant No. FDCT/060/2022/AFJ, National Natural Science Foundation of China Grant No. 62261160648), the UM and UMDF (Grant Nos. MYRG-GRG2023-00006-FST-UMDF, MYRG-GRG2024-00165-FST-UMDF, EF2024-00185-FST), and the National Natural Science Foundation of China (Grant No. 62266013).



\clearpage
\appendix
\label{sec:appendix}

\section{MQM Hierarchy}
\label{appendix_mqm_hierarchy}
The error typology used in {\framework} is categorized and further refined into 5 core error types and 19 distinct subtypes. Considering the characteristics and practical requirements of the proposed framework, specific error definitions within MQM have been further refined without altering their original meanings. These refinements are based on the MQM-\textit{Full Master File Instructions} and build upon the previous work~\cite{freitag-etal-2021-experts}, as detailed in Table~\ref{tab:MQM_error_definition}.

\section{MQM Error Weight}
\label{appendix_mqm_weight}
Following the architectural design and operational requirements of {\framework}, we eliminate the ``non-translation'' category and assign error weights that remain closely aligned with established human assessment practices~\cite{freitag-etal-2021-experts} for machine translation evaluation. The MQM error weights used for calculating the final score are shown in Table~\ref{tab:severity_weights}.

\section{Analysis on Token Consumption}
\label{appendix_average_token_consumption}
We present additional details on the average number of tokens consumed per sample in our experiments across various models as shown in Table~\ref{tab:average_token_consumption}. Multi-agent systems achieve performance improvements at a higher computational cost, which can be regarded as a form of test-time scaling~\citep{yang2025revisiting}. The token consumption per sample for M-MAD varies from 8.6k to 10.8k, while for HiMATE, it ranges from 12.6k to 14.4k. Although HiMATE exhibits higher token consumption compared to M-MAD, it concurrently delivers a substantial enhancement in performance.

\section{Stage-Transition Threshold Acquisition}
\label{appendix_threshold}
To determine an appropriate stage-transition threshold, several sentences unrelated to the primary experiment and not utilized as two-shot demonstrations are randomly selected from the MQM20 and MQM21 datasets, both for ZH-EN and EN-DE, forming the validation set. The size of this validation set is approximately one-tenth of the MQM22 dataset utilized in the main experiments.\footnote{The validation set can be found in \url{https://anonymous.4open.science/r/HiMATE-Anony}} Each sentence undergoes subtype evaluation and subsequent self-reflection to produce confidence scores. Subsequently, these confidence scores are ranked, and empirical analysis suggests the confidence score at the 60\% percentile as the optimal stage-transition threshold. Threshold values vary across language pairs and models.

\begin{table}[!t]
    \centering
    \caption{MQM error weights used to calculate the final score in {\framework}.}
    \begin{tabular}{@{}lll@{}}  
        \toprule
        Severity & Category & Weight \\ 
        \midrule
        Major & all others & 5 \\ 
        \midrule
        Minor & Fluency/Punctuation & 0.1 \\ 
              & all others & 1 \\ 
        \midrule
        Neutral & all & 0 \\ 
        \bottomrule
    \end{tabular}
    \label{tab:severity_weights}
\end{table}

\begin{table}[!t]
    \centering
    \caption{Average Token Consumption Per Sample. The abbreviations \textit{4o-mini} and \textit{qwen2.5} refer to the gpt-4o-mini and qwen2.5-72b-instruct models}
    \begin{tabular}{lccc}
        \toprule
        \textbf{Metric} & \textbf{4o-mini} & \textbf{qwen2.5} & \textbf{qwen-max} \\
        \midrule
        M-MAD  & 8,665  & 9,380  & 10,773 \\
        HiMATE & 12,600 & 15,927 & 14,408 \\
        \bottomrule
    \end{tabular}
    \label{tab:average_token_consumption}
\end{table}

\section{Matching Algorithm of Error Span Detection}
\label{Appendix_error_span}
A matching algorithm is employed to validate the accuracy of error span detection. Specifically, given a gold-standard error span $G$ and a detected error span $E$, the match is confirmed only if both overlap ratios satisfy $|I|/|G| \geq \theta$ and $|I|/|E| \geq \theta$, where $I$ denotes the intersection segment $G \cap E$, $|\cdot|$ represents length measurement~(e.g., measured by token or character count), and $\theta$ is a predefined matching threshold $0 < \theta \leq 1$. For instance, when $\theta = 50\%$, if the gold span $G$ is ``go back to the lab'' (4 tokens) and the detected span $E$ is ``back to the lab tomorrow'' (5 tokens), their intersection $I$ ``back to the lab'', 3 tokens) yields overlap ratios of 75\% (3/4 for $G$) and 60\% (3/5 for $E$), both exceeding the matching threshold 50\%. The pseudocode is shown as Algorithm~\ref{alg:fuzzy_match_while}.

\begin{algorithm}[t]
\relscale{0.95}
\linespread{1.2}\selectfont
\caption{Matching Algorithm}
\label{alg:fuzzy_match_while}
\begin{algorithmic}[1]
\State \textbf{Given:} Gold-standard error span $G$, detected error span $E$, matching threshold $\theta \in (0,1]$.
\State Tokenize $G \to G_t$, $E \to E_t$
\State $n \gets |G_t|$, $m \gets |E_t|$
\State $L \gets \min(n, m)$

\While{$L \geq 1$}
  \For{$i = 0$ \textbf{to} $n-L$}
    \If{$\exists\ j$ \text{s.t.} $G_t[i:i{+}L] = E_t[j:j{+}L]$}
      \State $\alpha \gets L/n$, $\beta \gets L/m$
      \If{$\alpha \geq \theta$ \textbf{and} $\beta \geq \theta$}
        \State \Return \textbf{true} 
      \EndIf
    \EndIf
  \EndFor
  \State $L \gets L - 1$
\EndWhile
\State \Return \textbf{false}
\end{algorithmic}
\end{algorithm}

\section{Supplementary Analysis on Domain-Specific Performance}
\label{appendix_cross_domain}
\begin{figure*}[!ht]
    \centering
    \vspace*{-13cm}
    \includegraphics[width=1\linewidth]{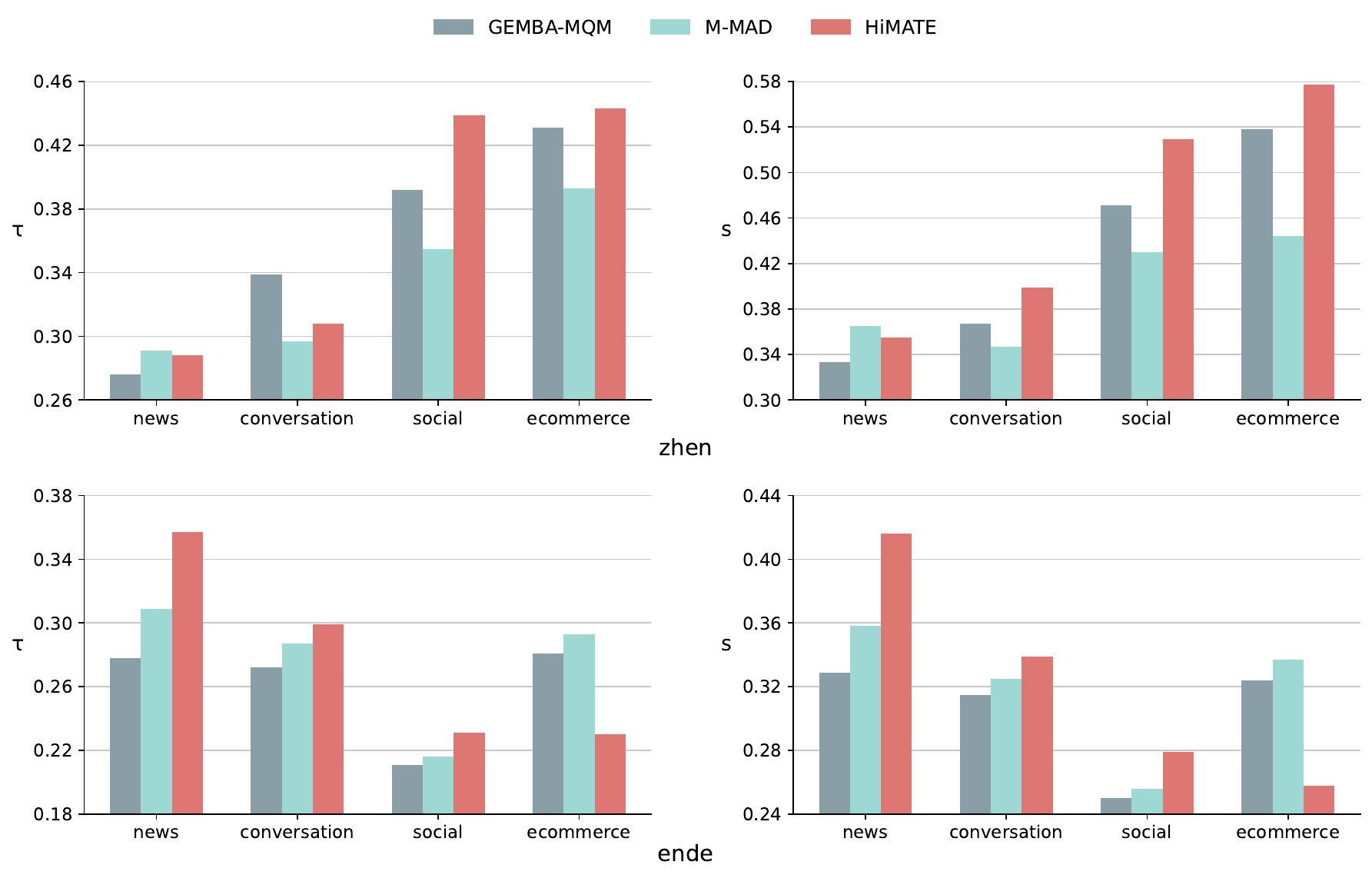}
    \caption{The Kendall's correlation coefficient $\tau$ and Spearman's correlation coefficient $s$ across multiple text domains on MQM22 ZH-EN and EN-DE dataset using the gpt-4o-mini model.}
    \label{fig:cross_domain_appendix}
\end{figure*}

As illustrated in Figure~\ref{fig:cross_domain_appendix}, although {\framework} may exhibit slightly lower performance in specific domains, it demonstrates strong alignment with manual evaluation across most scenarios. Incorporating a hierarchical multi-agent structure combined with fine-grained error detection ensures robust performance of {\framework} across diverse text domains.

\section{Prompts of {\framework} Agents}
\label{appendix_prompts}
The prompts utilized by the proposed {\framework} are presented in Tables~\ref{tab:Prompts_1} and~\ref{tab:Prompts_2}. A three-stage pipeline is adopted, where the self-reflection stage comprises two sequential steps, and the collaborative discussion stage involves a dialogue procedure between two tiers of agents.

\section{Detailed Case Example}
\label{appendix_case_study}
This section presents three illustrative cases demonstrating the complete workflow of each agent within {\framework}. These examples are generated utilizing the gpt-4o-mini model. The first example, sourced from MQM22 ZH-EN, corresponds to the case discussed in Chapter~\ref{case_study_main}, detailed in Tables~\ref{tab:first_case_1} and~\ref{tab:first_case_2}. The second example, also derived from MQM22 ZH-EN, is illustrated from Table~\ref{tab:second_case_1} to Table~\ref{tab:second_case_3}. The third example, originating from MQM22 EN-DE, is presented in Tables~\ref{tab:third_case_1} and~\ref{tab:third_case_2}.

\clearpage

\begin{table*}[t]
    \centering
    \vspace*{-2cm}
    \relscale{1.0}
    \renewcommand{\arraystretch}{1.2}
    \caption{Core error subtypes and descriptions after refinement. Here, 'Core' refers to high-level error categories at tier-1, while 'Subtype' denotes fine-grained error classifications at tier-2. 'Description' are utilized within the prompt to characterize each error type.}
    \begin{tabular*}{\textwidth}{@{\extracolsep{\fill}}p{1.8cm}p{3.9cm}|p{9.4cm}@{}}
        \toprule
        \textbf{Core} & \textbf{Subtype} & \textbf{Description} \\
        \hline
        \multirow{9}*{Accuracy}
        & \multirow{2}*{Addition} & Translation includes information (including the punctuation) not present in the source. \\
        & \multirow{3}*{Omission} & A paragraph present in the source (including the noun, verb, adverb, adverbial, punctuation, and so on) is missing in the translation. \\
        & \multirow{2}*{Mistranslation} & Incorrect use in target content of a word, inconsistent match, and the incorrect segmentation. \\
        & \multirow{2}*{Untranslated text} & Source text has been left untranslated. Not translating special symbols or placeholders is not an untranslated text error. \\
        \hline
        
        \multirow{3}*{\centering Terminology}
        & \multirow{2}*{Inappropriate for context} & Use of a term that differs from term usage required by a specified termbase or other resource. \\
        & Inconsistent use & Terminology is used inconsistently. \\
        \hline
        
        \multirow{11}*{Fluency}
        & \multirow{3}*{Punctuation} & Unpaired quote marks or parentheses. Missing mark from a set of paired punctuation marks, such as a missing parenthesis or quote mark. And the omission or addition of punctuation. \\
        & Spelling & Error occurring when a word is misspelled. \\
        & \multirow{2}*{Grammar} & Subject-verb disagreement, incorrect verb tenses or forms, and improper declension of nouns, pronouns, or adjectives. \\
        & \multirow{2}*{Register} & Wrong grammatical register (e.g. inappropriately informal pronouns). \\
        & Inconsistency & Internal inconsistency (not related to terminology). \\
        & \multirow{2}*{Character encoding} & Error occurring when characters are garbled due to incorrect application of an encoding. \\
        \hline
        \multirow{3}*{Style}
        & \multirow{3}*{Awkward} & Some languages provide grammatical and stylistic features that support complex, embedded ideas, which can result in awkward style if a target text retains these features. \\
        \hline
        \multirow{12}*{\parbox{2cm}{Locale \\[-0.5ex] Convention}}
        & \multirow{2}*{Address format} & Error involving inappropriate address format for locale. Such as the wrong form used for postal codes for target locale. \\
        & Currency format & Incorrect currency format for its locale. \\
        & Date format & Error involving inappropriate date format for its locale. \\
        & \multirow{3}*{Name format} & Name format used in text or a data field inappropriate for its locale, such as switching the order of last and first names inappropriately. \\
        & \multirow{2}*{Telephone format} & Error involving inappropriate telephone number form for locale. \\
        & \multirow{2}*{Time format} & Error involving incorrect time format for its locale. For example, unless specified as using a 24-hour clock, US time formats report time after noon using 12-hour notation. \\
        \bottomrule
    \end{tabular*}
    
    \label{tab:MQM_error_definition}
\end{table*}

\clearpage

\newpage
\clearpage

\begin{table*}[htbp]
    \centering
    \vspace*{-9cm}
    \caption{The prompt of Subtype Evaluation and Self-Reflection stages. Here, the Self-Reflection stage comprises two sequential steps, the error correction and comparative verification.}
    \begin{tabular*}{\textwidth}{@{\extracolsep{\fill}}p{5cm} p{5cm} p{5cm}@{}}
        \toprule
        \textit{Subtype Evaluation} & \multicolumn{2}{c}{\textit{Self-Reflection}} \\
        \midrule
        \textbf{Role-Play Prompt} & \textbf{Role-Play Prompt} & \textbf{Role-Play Prompt} \\
        You are an evaluator conducting a manual translation evaluation based on the MQM framework, and you need to determine whether the following translation from Chinese to English below contains an error. & 
        You are an evaluator conducting a manual translation evaluation based on the MQM framework. Here you are given a pair of sentences: a source Chinese sentence and its English translation. You need to correct the errors in the translation.  & 
        You are an evaluator conducting a manual translation evaluation based on the MQM framework. Here you need to compare the following original translation and the corrected translation based on the error definition. \\
         & & \\
        \textbf{Task Assignment Prompt} & \textbf{Task Assignment Prompt} & \textbf{Task Assignment Prompt} \\
        \textcolor{blue}{\{Source\} \newline \{Translation\} \newline \{Error Definition\}} \newline \newline \newline
        Please check if there are errors in the translation and whether they meet the definition of error types. You also need to indicate the degree of error. Error severities are assigned independently of category, and consist of Major, Minor, and Neutral levels, corresponding, respectively, to actual translation or grammatical errors, smaller imperfections, and purely subjective opinions about the translation. & 
        \textcolor{blue}{\{Source\} \newline \{Translation\} \newline \{Error Definition\} \newline \{Error Information\}} \newline \newline
        Previously, you have identified specific types of errors and the severity based on the error definition. Based on the information below, please correct the errors in the translation. & 
        \textcolor{blue}{\{Original Translation\} \newline \{Corrected Translation\}} \newline \newline \newline \newline
        Previously, you have corrected the translation. Now, determine whether the error mentioned previously in the original translation has been corrected. \\
        \bottomrule
    \end{tabular*}
    
    \label{tab:Prompts_1}
\end{table*}

\newpage
\clearpage

\begin{table*}[ht]
    \centering
    \vspace*{-11cm}
    \caption{The prompt of the Collaborative Discussion stage. Two tiers of agents participate in this stage and engage in discussion.}
    \begin{tabular*}{\textwidth}{@{\extracolsep{\fill}}p{7cm} p{7cm}@{}}
        \toprule
        \multicolumn{2}{c}{\textit{Collaborative Discussion}} \\
        \midrule
        \multicolumn{1}{c}{\textbf{Tier-1 Agent Round}} & \multicolumn{1}{c}{\textbf{Tier-2 Agent Round}} \\
        \midrule
        \textbf{Role-Play Prompt} & \textbf{Role-Play Prompt} \\
        You are an expert in manual translation evaluation based on the MQM framework and are responsible for a high-level error type. In previous work, an evaluator responsible for a subtype has expressed his lack of confidence in his answers. Please review his work and provide your opinions.\newline & 
        You are an evaluator who manually evaluates translations based on the MQM framework. In your previous work, you were not sure about the accuracy of your evaluation. Therefore, you communicated with the expert in charge of coordinating the evaluation of the error subtype for which you are responsible.\newline \\
        
        \textbf{Task Assignment Prompt} & \textbf{Task Assignment Prompt} \\
        \textcolor{blue}{\{Source Sentence\} \newline \{Translation\} \newline \{Previous Error Information\} \newline \{Previous Chat History\}}  \newline \newline
        The following is the source sentence and its corresponding translation, as well as the assessment of the subcategory evaluator on the existence and severity of the error. Please provide your opinion. You need to determine whether an error exists, and then judge whether the degree of error is correct. You don't need to completely negate it. & 
        \textcolor{blue}{\{Source Sentence\} \newline \{Translation\} \newline \{Previous Error Information\} \newline \{Previous Chat History\}} \newline \newline
        Regarding the source sentence and its translation, as well as the evaluation answer you provided, your superior evaluation expert has provided the following opinions. Please indicate whether you agree with his viewpoint. \\
        \bottomrule
    \end{tabular*}
    
    \label{tab:Prompts_2}
\end{table*}

\newpage
\clearpage

\begin{table*}[ht]
    \centering
    \vspace*{-6cm}
    \renewcommand{\arraystretch}{1.1}  
    \caption{The workflow for each agent in the first case from the MQM22 ZH-EN dataset generated by gpt-4o-mini. The table illustrates the subtype evaluation stage, with segments highlighted in \textcolor{red}{red} indicating incorrect judgments and those in \textcolor{green!50!black}{green} indicating correct ones. Subsequent tables will consistently adopt this convention.}
    \begin{tabular*}{\textwidth}{@{\extracolsep{\fill}}m{3.8cm} p{11.8cm}@{}}
        \toprule
        \multirow{1}{*}{\textbf{Source}} & 
        \begin{CJK*}{UTF8}{gbsn}希望你们了解一下
        \end{CJK*} \\
        \cdashline{1-2}
        \multirow{1}{*}{\textbf{Translation}} & 
        I hope you know about it. \\
        \cdashline{1-2}
        \multirow{1}{*}{\textbf{Reference}} & 
        I hope you can find out about it. \\
        \cdashline{1-2}
        \multirow{1}{*}{\textbf{Annotation}} & 
        Major-Mistranslation-‘know about it’ \\
        \midrule

        \multirow{31}{*}{{\framework}-SE} & 
        \textcolor{green!50!black}{Major}-\textcolor{green!50!black}{Mistranslation}-‘\textcolor{green!50!black}{know about it}’
        tier-2 agent: ``The translation ``I hope you know about it'' does not accurately convey the meaning of the source text ``\begin{CJK*}{UTF8}{gbsn}希望你们了解一下
        \end{CJK*}'' which more appropriately translates to ``I hope you all understand'' or ``I hope you all are aware'' The use of ``know about it'' implies a lesser degree of understanding and does not capture the intent of encouraging awareness or understanding, thus constituting a major mistranslation.''
        \newline  
        \textcolor{red}{Minor}-\textcolor{red}{Inappropriate for context}-‘\textcolor{green!50!black}{know}’
        tier-2 agent: ``The translation ``I hope you know about it'' does not accurately convey the nuance of ``\begin{CJK*}{UTF8}{gbsn}希望你们了解一下\end{CJK*},'' which suggests a desire for the audience to gain a better understanding or awareness of something, rather than simply knowing about it. The term ``\begin{CJK*}{UTF8}{gbsn}了解\end{CJK*}'' implies a deeper comprehension than ``know,'' which could be seen as inappropriate for the context. However, the overall meaning is still conveyed, hence the error is classified as minor.''
        \newline  
        \textcolor{red}{Major}-\textcolor{red}{Omission}-‘\textcolor{red}{\begin{CJK*}{UTF8}{gbsn}希望\end{CJK*}}’
        tier-2 agent: ``The translation omits the word ``\begin{CJK*}{UTF8}{gbsn}希望\end{CJK*}'' (``hope'') in the source, which changes the tone and intention of the message. The translation does not convey the speaker's hope or desire for understanding, resulting in a significant difference in meaning.''
        \newline  
        \textcolor{red}{Minor}-\textcolor{red}{Awkward}-‘I hope you \textcolor{green!50!black}{know about it}’
        tier-2 agent: ``The translation ``I hope you know about it'' is somewhat awkward because it implies a level of familiarity that may not be intended in the original Chinese phrase ``\begin{CJK*}{UTF8}{gbsn}希望你们了解一下
        \end{CJK*}'' which suggests a desire for the listeners to gain some understanding or awareness rather than simply knowing about something. A more natural translation could be ``I hope you can understand this'' or ``I hope you can learn about this'' which would better convey the intent of the original message.''
        \newline  
        \textcolor{red}{Minor}-\textcolor{red}{Addition}-‘\textcolor{green!50!black}{know about it}’
        tier-2 agent: ``The translation ``I hope you know about it'' is somewhat awkward because it implies a level of familiarity that may not be intended in the original Chinese phrase ``\begin{CJK*}{UTF8}{gbsn}希望你们了解一下
        \end{CJK*}'' which suggests a desire for the listeners to gain some understanding or awareness rather than simply knowing about something. A more natural translation could be ``I hope you can understand this'' or ``I hope you can learn about this'' which would better convey the intent of the original message.'' \\
        \bottomrule
    \end{tabular*}
    
    \label{tab:first_case_1}
\end{table*}

\newpage
\clearpage

\begin{table*}[ht]
    \centering
    \vspace*{-2cm}
    \renewcommand{\arraystretch}{1.1}  
    \caption{The workflow for each agent in the first case from the MQM22 ZH-EN dataset generated by gpt-4o-mini. The table illustrates the self-reflection and collaborative discussion stages.}
    \begin{tabular*}{\textwidth}{@{\extracolsep{\fill}}m{3.8cm} p{11.8cm}@{}}
        \toprule
        \multirow{1}{*}{\textbf{Source}} & 
        \begin{CJK*}{UTF8}{gbsn}希望你们了解一下
        \end{CJK*} \\
        \cdashline{1-2}
        \multirow{1}{*}{\textbf{Translation}} & 
        I hope you know about it. \\
        \cdashline{1-2}
        \multirow{1}{*}{\textbf{Reference}} & 
        I hope you can find out about it. \\
        \cdashline{1-2}
        \multirow{1}{*}{\textbf{Annotation}} & 
        Major-Mistranslation-‘know about it’ \\
        \midrule

        \multirow{8}{*}{{\framework}-SR-Correction} & 
        \textcolor{green!50!black}{Major}-\textcolor{green!50!black}{Mistranslation}-‘\textcolor{green!50!black}{know about it}’
        tier-2 agent: ``I hope you all understand.''
        \newline  
        \textcolor{red}{Minor}-\textcolor{red}{Inappropriate for context}-‘\textcolor{green!50!black}{know}’
        tier-2 agent: ``I hope you can understand it better.''
        \newline  
        \textcolor{red}{Major}-\textcolor{red}{Omission}-‘\textcolor{red}{\begin{CJK*}{UTF8}{gbsn}希望\end{CJK*}}’
        tier-2 agent: ``I hope you understand this.''
        \newline  
        \textcolor{red}{Minor}-\textcolor{red}{Awkward}-‘I hope you \textcolor{green!50!black}{know about it}’
        tier-2 agent: ``I hope you can understand this.'' 
        \newline  
        \textcolor{red}{Minor}-\textcolor{red}{Addition}-‘\textcolor{green!50!black}{know about it}’
        tier-2 agent: ``I hope you understand a bit about it.'' \\
        \cdashline{1-2}
        
        \multirow{8}{*}{{\framework}-SR-Verification} & 
        \textcolor{green!50!black}{Major}-\textcolor{green!50!black}{Mistranslation}-‘\textcolor{green!50!black}{know about it}’
        tier-2 agent: ``Error has been corrected.''
        \newline  
        \textcolor{red}{Minor}-\textcolor{red}{Inappropriate for context}-‘\textcolor{green!50!black}{know}’
        tier-2 agent: ``No significant difference between the original translation and the corrected Translation.''
        \newline  
        \textcolor{red}{Major}-\textcolor{red}{Omission}-‘\textcolor{red}{\begin{CJK*}{UTF8}{gbsn}希望\end{CJK*}}’
        tier-2 agent: ``No significant difference between the original translation and the corrected Translation.''
        \newline  
        \textcolor{red}{Minor}-\textcolor{red}{Awkward}-‘I hope you \textcolor{green!50!black}{know about it}’
        tier-2 agent: ``Error has been corrected.'' 
        \newline  
        \textcolor{red}{Minor}-\textcolor{red}{Addition}-‘\textcolor{green!50!black}{know about it}’
        tier-2 agent: ``Error has been corrected.'' \\
        \cdashline{1-2}
        
        \multirow{20}{*}{{\framework}-CD-Round1} & 
        \textcolor{green!50!black}{Major}-\textcolor{green!50!black}{Mistranslation}-‘\textcolor{green!50!black}{know about it}’
        tier-1 agent: ``Error Exist: Yes\quad Error Severity: Major.\quad The evaluator's assessment is correct. The translation ``I hope you know about it'' does not accurately capture the meaning of the source sentence ``\begin{CJK*}{UTF8}{gbsn}希望你们了解一下\end{CJK*}'' which conveys a sense of encouraging understanding or familiarity rather than merely hoping that the audience knows something. This misrepresentation of intent in the translation qualifies as a major error, as it alters the original message significantly.''
        \newline  
        \textcolor{red}{Minor}-\textcolor{red}{Awkward}-‘I hope you \textcolor{green!50!black}{know about it}’
        tier-1 agent: ``Error Exist: No.\quad Error Severity: Neutral.\quad The translation ``I hope you know about it'' is grammatically correct but does not fully capture the nuance of the source sentence ``\begin{CJK*}{UTF8}{gbsn}希望你们了解一下\end{CJK*}'' which implies a desire for understanding rather than mere knowledge. The awkwardness stems from the choice of words rather than a clear error, making it more subjective. Therefore, I would classify the severity as Neutral rather than Minor.'' 
        \newline  
        \textcolor{red}{Minor}-\textcolor{red}{Addition}-‘\textcolor{green!50!black}{know about it}’
        tier-1 agent: ``Error Exist: No.\quad Error Severity: Neutral.\quad The evaluator correctly identified the addition error, as the translation introduces a nuance that is not present in the source text. However, the severity should be considered neutral rather than minor, as the addition does not significantly distort the original meaning or create a major misunderstanding. It is more of a stylistic difference rather than a critical error.'' \\
        \cdashline{1-2}
        
        \multirow{3}{*}{{\framework}-CD-Round2} & 
        \textcolor{green!50!black}{Major}-\textcolor{green!50!black}{Mistranslation}-‘\textcolor{green!50!black}{know about it}’
        tier-2 agent: ``I agree with him.''
        \newline
        \textcolor{red}{Minor}-\textcolor{red}{Awkward}-‘I hope you \textcolor{green!50!black}{know about it}’
        tier-2 agent: ``I agree with him.'' 
        \newline  
        \textcolor{red}{Minor}-\textcolor{red}{Addition}-‘\textcolor{green!50!black}{know about it}’
        tier-2 agent: ``I agree with him.'' \\
        \cdashline{1-2}
        \bottomrule
    \end{tabular*}
    
    \label{tab:first_case_2}
\end{table*}

\begin{table*}[ht]
    \centering
    \vspace*{-4.5cm}
    \renewcommand{\arraystretch}{1.1}  
    \caption{The workflow for each agent in the second case from the MQM22 ZH-EN dataset generated by gpt-4o-mini. The table illustrates the subtype evaluation stage.}
    \begin{tabular*}{\textwidth}{@{\extracolsep{\fill}}m{3.8cm} p{11.8cm}@{}}
        \toprule
        \multirow{1}{*}{\textbf{Source}} & 
        \begin{CJK*}{UTF8}{gbsn}无论从接待、化妆、服装、灯光、摄影到后期都无可挑剔。
        \end{CJK*} \\
        \cdashline{1-2}
        \multirow{2}{*}{\textbf{Translation}} & 
        From reception, makeup, costumes, lighting, photography to the late stage, it is impeccable. \\
        \cdashline{1-2}
        \multirow{2}{*}{\textbf{Reference}} & 
        It was perfect in reception, makeup, clothing, lamplight, photography and post-editing. \\
        \cdashline{1-2}
        \multirow{1}{*}{\textbf{Annotation}} & 
        Major-Mistranslation-‘late stage’ \\
        \midrule

        \multirow{31}{*}{{\framework}-SE} & 
        \textcolor{green!50!black}{Major}-\textcolor{green!50!black}{Mistranslation}-‘the \textcolor{green!50!black}{late stage}’
        tier-2 agent: ``The phrase ``to the late stage'' in the translation does not accurately convey the meaning of the original Chinese text, which implies that everything is impeccable ``from reception to post-production'' without a clear distinction or segmentation of ``the late stage.'' This mistranslation leads to confusion about the intended meaning and scope of the description.''
        \newline  
        \textcolor{red}{Minor}-\textcolor{red}{Addition}-‘the \textcolor{green!50!black}{late stage}’
        tier-2 agent: ``The translation introduces the phrase ``the late stage'' which is not present in the source text (``\begin{CJK*}{UTF8}{gbsn}后期\end{CJK*}'' could refer to a general later stage rather than a specific term). This addition alters the original meaning slightly by being more specific than the original context intended.''
        \newline  
        \textcolor{red}{Major}-\textcolor{red}{Omission}-‘\textcolor{red}{\begin{CJK*}{UTF8}{gbsn}无论从\end{CJK*}}’
        tier-2 agent: ``The translation omits the phrase ``{\begin{CJK*}{UTF8}{gbsn}无论从\end{CJK*}}'' (which means ``regardless of'' or ``no matter if'') from the original Chinese sentence, altering the meaning and emphasis of the statement. This results in a significant loss of context and intent, qualifying it as a major omission.''
        \newline  
        \textcolor{red}{Minor}-\textcolor{red}{Grammar}-‘to the \textcolor{green!50!black}{late stage}’
        tier-2 agent: ``The phrase ``to the late stage'' does not accurately reflect the intended meaning of ``\begin{CJK*}{UTF8}{gbsn}后期\end{CJK*}'' which typically refers to ``post-production'' or ``editing'' in a context involving photography or film. Although the overall translation is understandable, this specific phrase could lead to ambiguity regarding the stages mentioned, making it a minor grammatical error.''
        \newline  
        \textcolor{red}{Minor}-\textcolor{red}{Inconsistency}-‘the \textcolor{green!50!black}{late stage}’
        tier-2 agent: ``The translation uses \``the late stage\'' which is not a commonly used term in English for ``\begin{CJK*}{UTF8}{gbsn}后期\end{CJK*}'' in this context. A more appropriate term would be ``post-production'' or ``final stage.'' This creates a slight inconsistency in the translation, but it does not significantly alter the meaning.''
        \newline  
        \textcolor{red}{Minor}-\textcolor{red}{Awkward}-‘to the \textcolor{green!50!black}{late stage}’
        tier-2 agent: ``The translation ``From reception, makeup, costumes, lighting, photography to the late stage, it is impeccable.'' feels awkward due to the phrasing ``to the late stage'' which is not a commonly used expression in English. A more natural phrasing like ``post-production'' would make it smoother. However, the overall meaning is conveyed, which keeps the severity at a minor level.'' \\
        \bottomrule
    \end{tabular*}
    
    \label{tab:second_case_1}
\end{table*}

\newpage
\clearpage

\begin{table*}[ht]
    \centering
    \vspace*{-10cm}
    \renewcommand{\arraystretch}{1.1}  
    \caption{The workflow for each agent in the second case from the MQM22 ZH-EN dataset generated by gpt-4o-mini. The table illustrates the self-reflection stage.}
    \begin{tabular*}{\textwidth}{@{\extracolsep{\fill}}m{3.8cm} p{11.8cm}@{}}
        \toprule
        \multirow{1}{*}{\textbf{Source}} & 
        \begin{CJK*}{UTF8}{gbsn}无论从接待、化妆、服装、灯光、摄影到后期都无可挑剔。
        \end{CJK*} \\
        \cdashline{1-2}
        \multirow{2}{*}{\textbf{Translation}} & 
        From reception, makeup, costumes, lighting, photography to the late stage, it is impeccable. \\
        \cdashline{1-2}
        \multirow{2}{*}{\textbf{Reference}} & 
        It was perfect in reception, makeup, clothing, lamplight, photography and post-editing. \\
        \cdashline{1-2}
        \multirow{1}{*}{\textbf{Annotation}} & 
        Major-Mistranslation-‘late stage’ \\
        \midrule

        \multirow{13}{*}{{\framework}-SR-Correction} & 
        \textcolor{green!50!black}{Major}-\textcolor{green!50!black}{Mistranslation}-‘the \textcolor{green!50!black}{late stage}’
        tier-2 agent: ``From reception, makeup, costumes, lighting, photography to post-production, everything is impeccable.''
        \newline  
        \textcolor{red}{Minor}-\textcolor{red}{Addition}-‘the \textcolor{green!50!black}{late stage}’
        tier-2 agent: ``From reception, makeup, costumes, lighting, photography to post-production, it is impeccable.''
        \newline  
        \textcolor{red}{Major}-\textcolor{red}{Omission}-‘\textcolor{red}{\begin{CJK*}{UTF8}{gbsn}无论从\end{CJK*}}’
        tier-2 agent: ``Regardless of reception, makeup, costumes, lighting, photography, or the late stage, it is impeccable.''
        \newline  
        \textcolor{red}{Minor}-\textcolor{red}{Grammar}-‘to the \textcolor{green!50!black}{late stage}’
        tier-2 agent: ``From reception, makeup, costumes, lighting, photography to post-production, it is impeccable.''
        \newline  
        \textcolor{red}{Minor}-\textcolor{red}{Inconsistency}-‘the \textcolor{green!50!black}{late stage}’
        tier-2 agent: ``From reception, makeup, costumes, lighting, photography to post-production, it is impeccable.''
        \newline  
        \textcolor{red}{Minor}-\textcolor{red}{Awkward}-‘to the \textcolor{green!50!black}{late stage}’
        tier-2 agent: ``From reception, makeup, costumes, lighting, photography to post-production, it is impeccable.'' \\
        \cdashline{1-2}
        
        \multirow{8}{*}{{\framework}-SR-Verification} & 
        \textcolor{green!50!black}{Major}-\textcolor{green!50!black}{Mistranslation}-‘the \textcolor{green!50!black}{late stage}’
        tier-2 agent: ``Error has been corrected.''
        \newline  
        \textcolor{red}{Minor}-\textcolor{red}{Addition}-‘the \textcolor{green!50!black}{late stage}’
        tier-2 agent: ``Error has been corrected.''
        \newline  
        \textcolor{red}{Major}-\textcolor{red}{Omission}-‘\textcolor{red}{\begin{CJK*}{UTF8}{gbsn}无论从\end{CJK*}}’
        tier-2 agent: ``No significant difference between the original translation and the corrected Translation.''
        \newline  
        \textcolor{red}{Minor}-\textcolor{red}{Grammar}-‘to the \textcolor{green!50!black}{late stage}’
        tier-2 agent: ``No significant difference between the original translation and the corrected Translation.''
        \newline  
        \textcolor{red}{Minor}-\textcolor{red}{Inconsistency}-‘the \textcolor{green!50!black}{late stage}’
        tier-2 agent: ``Error has been corrected.''
        \newline  
        \textcolor{red}{Minor}-\textcolor{red}{Awkward}-‘to the \textcolor{green!50!black}{late stage}’
        tier-2 agent: ``Error has been corrected.'' \\
        
        \bottomrule
    \end{tabular*}
    
    \label{tab:second_case_2}
\end{table*}

\newpage
\clearpage

\begin{table*}[ht]
    \centering
    \vspace*{-5cm}
    \renewcommand{\arraystretch}{1.1}  
    \caption{The workflow for each agent in the second case from the MQM22 ZH-EN dataset generated by gpt-4o-mini. The table illustrates the collaborative discussion stage.}
    \begin{tabular*}{\textwidth}{@{\extracolsep{\fill}}m{3.8cm} p{11.8cm}@{}}
        \toprule
        \multirow{1}{*}{\textbf{Source}} & 
        \begin{CJK*}{UTF8}{gbsn}无论从接待、化妆、服装、灯光、摄影到后期都无可挑剔。
        \end{CJK*} \\
        \cdashline{1-2}
        \multirow{2}{*}{\textbf{Translation}} & 
        From reception, makeup, costumes, lighting, photography to the late stage, it is impeccable. \\
        \cdashline{1-2}
        \multirow{2}{*}{\textbf{Reference}} & 
        It was perfect in reception, makeup, clothing, lamplight, photography and post-editing. \\
        \cdashline{1-2}
        \multirow{1}{*}{\textbf{Annotation}} & 
        Major-Mistranslation-‘late stage’ \\
        \midrule

        \multirow{38}{*}{{\framework}-CD-Round1} & 
        \textcolor{green!50!black}{Major}-\textcolor{green!50!black}{Mistranslation}-‘the \textcolor{green!50!black}{late stage}’
        tier-1 agent: ``Error Exist: Yes.\quad Error Severity: Major.\quad The translation ``From reception, makeup, costumes, lighting, photography to the late stage, it is impeccable'' misrepresents the original meaning. The term ``\begin{CJK*}{UTF8}{gbsn}后期\end{CJK*}'' typically refers to ``post-production'' rather than ``the late stage.'' This distortion significantly alters the intended message about the quality of the various aspects involved, thus justifying a major severity rating.''
        \newline  
        \textcolor{red}{Minor}-\textcolor{red}{Addition}-‘the \textcolor{green!50!black}{late stage}’
        tier-1 agent: ``Error Exist: No.\quad Error Severity: Neutral.\quad The evaluator correctly identified the addition error, as the phrase ``late stage'' in the translation does not correspond to the source text. However, the severity should be considered neutral rather than minor, as this error does not belong to an addition error. ''
        \newline  
        \textcolor{red}{Minor}-\textcolor{red}{Inconsistency}-‘the \textcolor{green!50!black}{late stage}’
        tier-1 agent: ``Error Exist: No.\quad Error Severity: Neutral.\quad The evaluator correctly identified the existence of an inconsistency in the translation, as the phrase ``late stage'' does not accurately reflect the original meaning of ``\begin{CJK*}{UTF8}{gbsn}后期\end{CJK*}'' which typically refers to ``post-production'' in this context. However, the severity should be considered Neutral rather than Minor, as this does not significantly impact the overall understanding of the text. The translation remains comprehensible, and the error does not lead to a misunderstanding of the main message.''
        \newline  
        \textcolor{red}{Minor}-\textcolor{red}{Awkward}-‘to the \textcolor{green!50!black}{late stage}’
        tier-2 agent: ``Error Exist: No.\quad Error Severity: Neutral.\quad The translation retains grammatical acceptability but lacks the fluidity and coherence expected in the target language. While it does exhibit awkwardness, the severity should be assessed as neutral rather than minor, as it primarily involves a subjective opinion rather than causing a notable impact on the overall meaning or readability. The phrase ``to the late stage'' feels particularly awkward and could confuse readers, but it does not constitute a significant error affecting comprehension.'' \\
        \cdashline{1-2}
        
        \multirow{4}{*}{{\framework}-CD-Round2} & 
        \textcolor{green!50!black}{Major}-\textcolor{green!50!black}{Mistranslation}-‘the \textcolor{green!50!black}{late stage}’
        tier-2 agent: ``I agree with him.''
        \newline  
        \textcolor{red}{Minor}-\textcolor{red}{Addition}-‘the \textcolor{green!50!black}{late stage}’
        tier-1 agent: ``I agree with him.''
        \newline  
        \textcolor{red}{Minor}-\textcolor{red}{Inconsistency}-‘the \textcolor{green!50!black}{late stage}’
        tier-1 agent: ``I agree with him.''
        \newline  
        \textcolor{red}{Minor}-\textcolor{red}{Awkward}-‘to the \textcolor{green!50!black}{late stage}’
        tier-2 agent: ``I agree with him.'' \\
        \cdashline{1-2}
        \bottomrule
    \end{tabular*}
    
    \label{tab:second_case_3}
\end{table*}

\newpage
\clearpage

\begin{table*}[ht]
    \centering
    \vspace*{-8cm}
    \renewcommand{\arraystretch}{1.1}  
    \caption{The workflow for each agent in the third case from the MQM22 EN-DE dataset generated by gpt-4o-mini. The table illustrates the subtype evaluation stage.}
    \begin{tabular*}{\textwidth}{@{\extracolsep{\fill}}m{3.8cm} p{11.8cm}@{}}
        \toprule
        \multirow{4}{*}{\textbf{Source}} & This move is just the latest twist in the Dallas-Washington rivalry, one that got juiced up even more this week when Cowboys head coach Mike McCarthy predicted a win for his team, prompting some fireworks back and forth with Washington's Ron Rivera and players. \\
        \cdashline{1-2}
        \multirow{4}{*}{\textbf{Translation}} & Dieser Schritt ist nur die jüngste Wendung in der Rivalität zwischen Dallas und Washington, die diese Woche noch stärker gemacht wurde, als Cowboys-Cheftrainer Mike McCarthy einen Sieg für sein Team vorhersagte und ein Feuerwerk zwischen Ron Rivera und Spielern aus Washington auslöste. \\
        \cdashline{1-2}
        \multirow{4}{*}{\textbf{Reference}} & Diese Aktion ist nichts als ein weiteres Kapitel in der Rivalität zwischen Dallas und Washington, die diese Woche noch mehr an Fahrt aufnahm, als der leitende Coach der Cowboys Mike McCarthy einen Sieg seines Teams voraussagte und ein hitziges Wortgefecht mit Ron Rivera von Washington und seinen Spielern auslöste. \\
        \cdashline{1-2}
        \multirow{2}{*}{\textbf{Annotation}} & 
        Minor-Awkward-‘stärker gemacht’, Minor-Awkward-‘Feuerwerk’, Minor-Mistranslation-‘Feuerwerk’ \\
        \midrule

        \multirow{17}{*}{{\framework}-SE} & 
        \textcolor{green!50!black}{Minor}-\textcolor{green!50!black}{Awkward}-‘\textcolor{green!50!black}{stärker gemacht} wurde’
        tier-2 agent: ``The phrase ``stärker gemacht wurde'' is somewhat awkward in German. A more natural phrasing could be ``intensiver wurde'' or ``mehr an Intensität gewann.'' The current translation retains the original structure but does not flow as smoothly in German, leading to a minor awkwardness in style.''
        \newline  
        \textcolor{red}{Major}-\textcolor{green!50!black}{Mistranslation}-‘\textcolor{green!50!black}{Feuerwerk}’
        tier-2 agent: ``The translation uses ``Feuerwerk'' which is more commonly associated with fireworks in a literal sense, while the context implies a figurative sense of heated exchanges or confrontations. A more appropriate term might be ``Auseinandersetzungen'' or ``Konflikte'' to convey the intended meaning of verbal disputes.''
        \newline  
        \textcolor{red}{Minor}-\textcolor{red}{Inconsistency}-‘die diese Woche noch \textcolor{green!50!black}{stärker gemacht} wurde’
        tier-2 agent: ``The phrase ``die diese Woche noch stärker gemacht wurde'' implies that the rivalry was made stronger, whereas the original phrase suggests a progression in intensity due to recent events. A more accurate translation could clarify that the rivalry was intensified rather than simply ``made stronger.'' This is a minor inconsistency that doesn't convey the precise meaning of the source text.'' \\
        
        \bottomrule
    \end{tabular*}
    
    \label{tab:third_case_1}
\end{table*}

\newpage
\clearpage

\begin{table*}[ht]
    \centering
    \vspace*{-0cm}
    \renewcommand{\arraystretch}{1.1}  
    \caption{The workflow for each agent in the third case from the MQM22 EN-DE dataset generated by gpt-4o-mini. The table illustrates the self-refleciton and collaborative discussion stages.}
    \begin{tabular*}{\textwidth}{@{\extracolsep{\fill}}m{3.8cm} p{11.8cm}@{}}
        \toprule
        \multirow{4}{*}{\textbf{Source}} & This move is just the latest twist in the Dallas-Washington rivalry, one that got juiced up even more this week when Cowboys head coach Mike McCarthy predicted a win for his team, prompting some fireworks back and forth with Washington's Ron Rivera and players. \\
        \cdashline{1-2}
        \multirow{4}{*}{\textbf{Translation}} & Dieser Schritt ist nur die jüngste Wendung in der Rivalität zwischen Dallas und Washington, die diese Woche noch stärker gemacht wurde, als Cowboys-Cheftrainer Mike McCarthy einen Sieg für sein Team vorhersagte und ein Feuerwerk zwischen Ron Rivera und Spielern aus Washington auslöste. \\
        \cdashline{1-2}
        \multirow{4}{*}{\textbf{Reference}} & Diese Aktion ist nichts als ein weiteres Kapitel in der Rivalität zwischen Dallas und Washington, die diese Woche noch mehr an Fahrt aufnahm, als der leitende Coach der Cowboys Mike McCarthy einen Sieg seines Teams voraussagte und ein hitziges Wortgefecht mit Ron Rivera von Washington und seinen Spielern auslöste. \\
        \cdashline{1-2}
        \multirow{2}{*}{\textbf{Annotation}} & 
        Minor-Awkward-‘stärker gemacht’, Minor-Awkward-‘Feuerwerk’, Minor-Mistranslation-‘Feuerwerk’ \\
        \midrule

        \multirow{15}{*}{{\framework}-SR-Correction} & 
        \textcolor{green!50!black}{Minor}-\textcolor{green!50!black}{Awkward}-‘\textcolor{green!50!black}{stärker gemacht} wurde’
        tier-2 agent: ``Dieser Schritt ist nur die jüngste Wendung in der Rivalität zwischen Dallas und Washington, die diese Woche noch intensiver wurde, als Cowboys-Cheftrainer Mike McCarthy einen Sieg für sein Team vorhersagte und ein Feuerwerk zwischen Ron Rivera und Spielern aus Washington auslöste.''
        \newline  
        \textcolor{red}{Major}-\textcolor{green!50!black}{Mistranslation}-‘\textcolor{green!50!black}{Feuerwerk}’
        tier-2 agent: ``Dieser Schritt ist nur die jüngste Wendung in der Rivalität zwischen Dallas und Washington, die diese Woche noch stärker gemacht wurde, als Cowboys-Cheftrainer Mike McCarthy einen Sieg für sein Team vorhersagte und Auseinandersetzungen zwischen Ron Rivera und Spielern aus Washington auslöste.''
        \newline  
        \textcolor{red}{Minor}-\textcolor{red}{Inconsistency}-‘die diese Woche noch \textcolor{green!50!black}{stärker gemacht} wurde’
        tier-2 agent: ``Dieser Schritt ist nur die jüngste Wendung in der Rivalität zwischen Dallas und Washington, die diese Woche noch intensiver wurde, als Cowboys-Cheftrainer Mike McCarthy einen Sieg für sein Team vorhersagte und ein Feuerwerk zwischen Ron Rivera und Spielern aus Washington auslöste.'' \\
        \cdashline{1-2}
        
        \multirow{6}{*}{{\framework}-SR-Verification} & 
        \textcolor{green!50!black}{Minor}-\textcolor{green!50!black}{Awkward}-‘\textcolor{green!50!black}{stärker gemacht} wurde’
        tier-2 agent: ``Error has been corrected.''
        \newline  
        \textcolor{red}{Major}-\textcolor{green!50!black}{Mistranslation}-‘\textcolor{green!50!black}{Feuerwerk}’
        tier-2 agent: ``Error has been corrected.''
        \newline  
        \textcolor{red}{Minor}-\textcolor{red}{Inconsistency}-‘die diese Woche noch \textcolor{green!50!black}{stärker gemacht} wurde’
        tier-2 agent: ``No significant difference between the original translation and the corrected Translation.'' \\
        \cdashline{1-2}
        
        \multirow{10}{*}{{\framework}-CD-Round1} & 
        \textcolor{green!50!black}{Minor}-\textcolor{green!50!black}{Awkward}-‘\textcolor{green!50!black}{stärker gemacht} wurde’
        tier-1 agent: ``No explanation due to exceeding the stage-transition threshold.''
        \newline  
        \textcolor{green!50!black}{Minor}-\textcolor{green!50!black}{Mistranslation}-‘\textcolor{green!50!black}{Feuerwerk}’
        tier-1 agent: ``Error Exist: Yes.\quad Error Severity: Minor.\quad While there is a mistranslation present, the errors noted do not significantly alter the overall meaning of the source text. The translation captures the core sentiment and information though with some imprecision. Therefore, although there is an error, it does not warrant a major severity classification, as it does not impede comprehension or alter the message's intent in a substantial way. A minor severity classification is more appropriate here.'' \\
        \cdashline{1-2}
        
        \multirow{1}{*}{{\framework}-CD-Round2} & 
        \textcolor{green!50!black}{Minor}-\textcolor{green!50!black}{Mistranslation}-‘\textcolor{green!50!black}{Feuerwerk}’
        tier-2 agent: ``I agree with him.'' \\
        \cdashline{1-2}
        \bottomrule
    \end{tabular*}
    
    \label{tab:third_case_2}
\end{table*}

\end{document}